\documentclass[sigconf]{acmart}
\AtBeginDocument{%
  }

\setcopyright{acmlicensed}
\copyrightyear{2018}
\acmYear{2018}
\acmDOI{XXXXXXX.XXXXXXX}
\acmConference[Conference acronym 'XX]{Make sure to enter the correct
  conference title from your rights confirmation email}{June 03--05,
  2018}{Woodstock, NY}
\acmISBN{978-1-4503-XXXX-X/2018/06}

\usepackage{amsmath}
\usepackage{amsthm}
\usepackage{booktabs}
\usepackage{framed} 
\usepackage{multirow}
\usepackage{enumitem}
\usepackage{caption}
\usepackage{graphicx}
\usepackage{float} 
\usepackage{etoolbox}
\usepackage{xspace}
\usepackage{tabularx}
\usepackage{tcolorbox}
\tcbuselibrary{skins,breakable}
\usepackage{accents}
\usepackage{subfig}
\usepackage{makecell}
\usepackage{bbding}
\usepackage{multicol}
\usepackage[table]{xcolor}
\usepackage{hyperref}
\usepackage[normalem]{ulem}

\newcommand{\eg}{\emph{e.g.},\xspace}

\newcommand{\model}{TimeWorld}
\newcommand{\bench}{TSWM-Bench}

\newcommand\figref[1]{Figure~\ref{#1}}

\newcommand\tabref[1]{Table~\ref{#1}}
\newcommand\secref[1]{Section~\ref{#1}}

\newcommand\appref[1]{Appendix~\ref{#1}}

\newcommand{\eat}[1]{}

\begin{document}

\title{\model: Towards Action-Conditioned Multimodal World Models for Time Series Forecasting}

\author{Hang Ni}
\orcid{0009-0002-1341-1602}
\affiliation{%
  \institution{The Hong Kong University of Science and Technology (Guangzhou)}
  \city{Guangzhou}
  \country{China}}
\email{hni017@connect.hkust-gz.edu.cn}

\author{Weijia Zhang}
\correspondingauthor
\orcid{0000-0001-5085-5216}
\affiliation{%
  \institution{The Hong Kong University of Science and Technology (Guangzhou)}
  \city{Guangzhou}
  \country{China}
  }
\email{wzhang411@connect.hkust-gz.edu.cn}

\author{Hao Liu}
\correspondingauthor
\orcid{0000-0003-4271-1567}
\affiliation{%
  \institution{The Hong Kong University of Science and Technology (Guangzhou)}
  \city{Guangzhou}
  \country{China}
  }
\email{liuh@ust.hk}

\renewcommand{\shortauthors}{Hang Ni, Weijia Zhang, \& Hao Liu}

\begin{abstract}
Time series forecasting supports decision-making in web applications such as online retail and network management, where planned actions can reshape future dynamics.
World models provide a natural paradigm for predicting such action-dependent changes beyond extrapolating historical patterns.
However, existing time series world models (TSWMs) typically rely on fixed, domain-specific covariate schemas that limit flexible intervention specification, while multimodal forecasters treat text primarily as passive predictive context.
To investigate these gaps, we first introduce \textbf{\bench}, a cross-domain benchmark for \emph{multimodal TSWMs} spanning seven real-world datasets, with textual interventions grounded in recorded action--outcome data.
Our evaluation reveals that conventional multimodal forecasters underperform history-only time series foundation models (TSFMs) despite dataset-specific training, whereas simple text-prefix adaptations improve forecasting across all evaluated TSFM backbones.
These findings suggest that pretrained TSFMs encode implicit world knowledge of temporal dynamics, offering a strong foundation for action-conditioned modeling beyond conventional multimodal fusion. 
Building on this insight, we further propose \textbf{\model}, a plug-in framework that adapts pretrained TSFMs through explicit action-response modeling.
Specifically, the Prototype-Guided Latent Modulation (PLM) module grounds textual actions in temporal dynamics by composing representative response prototypes to modulate future representations based on the action and historical patterns.
The Action Contrastive Learning (ACL) component strengthens action-specific response modeling by contrasting these updates under alternative interventions sharing the same history, distinguishing action types while preserving relative intensity relationships.
Experiments across diverse domains and TSFM backbones demonstrate consistent improvements over strong baselines.
\end{abstract}




\maketitle

\section{Introduction}
\label{sec:intro}

Time series forecasting (TSF) underpins web-related applications such as retail planning~\cite{salinas2020deepar}, network management~\cite{lazaris2019lstm}, and financial analytics~\cite{zhang2019deeplob}.
Recent time series foundation models (TSFMs) have further improved forecasting by learning transferable priors over temporal dynamics through large-scale pretraining~\cite{das2024decoder}.
However, forecasting in these applications often requires anticipating how planned actions alter future trajectories, beyond extrapolating historical patterns~\cite{cedar2026}.
For example, an online retailer often needs to estimate demand under alternative discount campaigns before allocating its promotional budget.
Motivated by the world-model paradigm of predicting system evolution under different actions~\cite{hafner2025mastering,genie2024}, we study \emph{action-conditioned time series world models} (TSWMs) that forecast future trajectories from historical observations and a specified action~\cite{zhang2026on}.

Existing TSWMs commonly encode interventions as numerical or categorical covariates~\cite{cedar2026,mobiwm2026,zhang2026on}.
Such predefined, domain-specific schemas struggle to capture complex intervention semantics (\eg buy-one-get-one offers), limiting cross-domain generalization~\cite{qwenrobot-world}.
Moreover, their reliance on stepwise future controls makes it difficult to express high-level temporal intervention patterns (\eg phasing out discounts) without prescribing exact control sequences.
Instead, natural language provides a flexible interface to specify action semantics, intensity, and temporal scope without fixed schemas or stepwise controls~\cite{pmlr-v305-jang25a}.
However, recent studies of multimodal time series forecasting typically use text to describe temporal patterns~\cite{timecma2025,dualsg2025}, provide domain metadata~\cite{timellm2024,unitime2024}, or supply contextual signals such as news and events~\cite{timemmd2024,wang2024from,cik2025,aurora2026}.
In these methods, text provides predictive context rather than specifying interventions whose effects must be modeled.
We therefore investigate \emph{multimodal TSWMs} that condition future temporal dynamics on textual actions.

Early efforts in multimodal TSWMs rely on synthetic data for counterfactual ground truth~\cite{tadiff2026} or limit evaluation to a single domain~\cite{reasoncast2026}.
To address these gaps, we first introduce \textbf{\bench}, a benchmark for multimodal TSWMs spanning seven real-world datasets across economic, physical, and healthcare domains, with interventions ranging from price adjustments to medical treatments.
Each instance pairs a univariate history and its observed future with textual action and context descriptions derived from source records.
\tabref{tab:benchmark-coverage} compares its coverage with existing benchmarks.
We evaluate history-only TSFMs, representative multimodal forecasters, and covariate-aware TSWMs to examine their capabilities for action-conditioned prediction.
Our results reveal that existing multimodal forecasting methods consistently underperform history-only TSFMs despite dataset-specific training.
By contrast, simple text-prefix adaptations of TSFMs improve forecasting accuracy across all evaluated backbones and even outperform covariate-aware TSWMs.
These findings motivate a new route to world modeling: \emph{reusing pretrained world-dynamics priors while learning how textual actions reshape future states}.

However, effectively adapting pretrained TSFMs to multimodal TSWMs presents two critical challenges:
\emph{(1) Action Grounding.}
Incorporating textual actions requires explicitly translating intervention semantics into state-dependent changes to the predictive dynamics encoded by pretrained TSFMs.
Existing multimodal forecasters typically fuse textual and temporal representations directly, complicating cross-modal interpretation~\cite{nimadi2026} while leaving fine-grained action effects on local temporal patterns implicit.
Consequently, capturing meaningful intervention responses and their state-dependent evolution remains challenging.
\emph{(2) Action Discrimination.}
Even with action grounding, forecasts may remain dominated by pretrained dynamics, obscuring action-specific responses.
Forecasting loss alone provides limited supervision for action--response alignment, as it evaluates factual outcomes without comparing responses to alternative actions under the same history.
This requires additional supervision for action discrimination to ensure that predictions faithfully reflect the specified intervention.

To address these challenges, we further propose \textbf{\model}, a plug-in framework for building action-conditioned time series world models from pretrained TSFMs.
For action grounding, \emph{Prototype-Guided Latent Modulation (PLM)} constructs a bank of representative temporal response prototypes and adaptively composes them according to the specified action, forecast interval, and temporal state.
The resulting responses modulate future hidden states within a compact latent subspace, preserving the pretrained temporal components.
For action discrimination, \emph{Action Contrastive Learning (ACL)} uses large language models (LLMs) to construct alternative interventions under the same history and applies contrastive supervision to their latent updates. This strengthens action--response alignment while preserving relative intensity relationships.
\emph{Context-enhanced Temporal Modeling (CTM)} further incorporates textual system context through lightweight prefix conditioning.
Extensive experiments demonstrate the effectiveness and generality of \model\xspace across diverse domains and TSFM backbones.

Our main contributions are threefold:
(1) We introduce \bench, 
the first cross-domain benchmark for multimodal TSWMs with textual interventions grounded in real-world action–outcome records.
(2) We propose \model, a general framework that augments pretrained TSFM dynamics through PLM-based action grounding and ACL-based discrimination.
(3) Extensive experiments demonstrate consistent improvements over strong baselines across diverse domains and TSFM backbones.

\begin{table}[t]
\centering
\caption{Comparison of benchmarks.}
\label{tab:benchmark-coverage}
\small
\resizebox{\linewidth}{!}{%
\begin{tabular}{l|cccc}
\toprule
\textbf{Study} & \makecell{\textbf{Action-}\\\textbf{Conditioned}} & \makecell{\textbf{Action}\\\textbf{Format}} & \makecell{\textbf{Real-world}\\\textbf{Action}} & \makecell{\textbf{Multi-}\\\textbf{domain}} \\
\midrule
\multicolumn{5}{l}{\emph{Multimodal TSF}} \\
Time-MMD~\cite{timemmd2024} & \XSolidBrush & -- & -- & \CheckmarkBold \\
CiK~\cite{cik2025} & \XSolidBrush & -- & -- & \CheckmarkBold \\
\midrule
\multicolumn{5}{l}{\emph{Structured TSWM}} \\
MobiWM~\cite{mobiwm2026} & \CheckmarkBold & Covariates & \XSolidBrush & \XSolidBrush \\
CEDAR~\cite{cedar2026} & \CheckmarkBold & Covariates & \CheckmarkBold & \XSolidBrush \\
Zhang et al.~\cite{zhang2026on} & \CheckmarkBold & Covariates & \CheckmarkBold & \CheckmarkBold \\
\midrule
\multicolumn{5}{l}{\emph{Multimodal TSWM}} \\
TADiff~\cite{tadiff2026} & \CheckmarkBold & Text & \XSolidBrush & \CheckmarkBold \\
ReasonCast~\cite{reasoncast2026} & \CheckmarkBold & Text & \CheckmarkBold & \XSolidBrush \\
\midrule
\textbf{\bench} & \CheckmarkBold & Text & \CheckmarkBold & \CheckmarkBold \\
\bottomrule
\end{tabular}%
}
\end{table}

\section{Related Work}
\label{sec:related_work}
In this section, we review time series foundation models, multimodal forecasting, and time series world models.

\noindent\textbf{Time Series Foundation Models.}
Large-scale pretraining enables TSFMs to acquire transferable world knowledge of temporal evolution across datasets~\cite{das2024decoder,niextreme2026}.
Existing approaches primarily follow two forecasting paradigms.
\emph{Autoregressive models} generate future patches sequentially, allowing flexible prediction horizons but potentially accumulating errors over long rollouts~\cite{timer2024,timemoe2025,moirai22025}.
\emph{Masked-future models} instead predict future patches jointly from historical observations and masked placeholders, enabling parallel inference without recursive error propagation~\cite{timesfm32026,chronos22025,toto22026}.
Despite this implicit knowledge of system dynamics, these models lack mechanisms to ground textual actions in state-dependent changes to future trajectories.

\noindent\textbf{Multimodal Time Series Forecasting.}
Multimodal forecasting enriches numerical observations with textual context, exploiting complementary semantic information to improve predictive accuracy.
Text serves varied roles: early efforts use \emph{task descriptions} to formulate forecasting as language generation~\cite{promptcast2024}, summarize \emph{temporal patterns} to expose numerical structure~\cite{timecma2025,dualsg2025}, or provide \emph{domain metadata} to contextualize heterogeneous series~\cite{timellm2024,unitime2024,timevlm2025}.
Beyond describing the task or series, \emph{predictive context} supplies supplementary information correlated with future observations but unavailable in numerical histories, including retrospective reports and news~\cite{timemmd2024,wang2024from}, event announcements~\cite{camef2025}, and prospective conditions or constraints~\cite{cik2025,tadiff2026}.
Multimodal foundation models further learn transferable numerical--textual representations through shared pretraining, extending context-aware forecasting across datasets~\cite{chattime2025,aurora2026,pythia2026,qiyaom2026}.
However, predominantly using text as passive predictive context leaves the causal effects of actively planned interventions insufficiently modeled, motivating the shift toward action-conditioned simulation.

\noindent\textbf{Time Series World Models.}
World models for control~\cite{hafner2025mastering} and interactive video generation~\cite{genie2024} motivate modeling temporal evolution for prediction and planning.
Inspired by video joint-embedding prediction~\cite{vjepa2024}, \emph{latent predictive modeling} forecasts future representations before decoding observations, but does not condition state transitions on actions~\cite{timecapsule2025,latenttsf2026}.
Although multimodal extensions incorporate contextual information~\cite{pythia2026,worldts2026}, they still do not explicitly model how controllable actions alter temporal state transitions.
In contrast, early \emph{action-conditioned} approaches use structured treatments for counterfactual prediction or clinical policy learning~\cite{ct2022,mambacdsp2025,meddreamer2026}.
Recent TSWMs encode actions as numerical or categorical covariates: base-station power and antenna settings~\cite{mobiwm2026}, advertising budgets~\cite{cedar2026}, or engineering controls and treatment doses~\cite{zhang2026on}.
These formulations restrict actions to domain-specific schemas and may omit intervention details.
Language offers a flexible interface without prescribing a unique control sequence, motivating \emph{text-conditioned multimodal TSWMs}.
For example, TADiff~\cite{tadiff2026} forecasts under alternative textual conditions through text-attributive diffusion, but its real-data evaluation concerns descriptive conditions and semantic consistency rather than recorded action effects, with ground-truth counterfactual trajectories confined to synthetic data.
ReasonCast~\cite{reasoncast2026} translates textual actions into semantic interventions through agentic reasoning, but focuses on e-commerce demand.
A comprehensive multimodal TSWM benchmark grounded in real action--outcome records across domains remains missing.
We fill this gap with \bench\xspace and introduce \model, which extends pretrained predictive dynamics to textual interventions through prototype-guided latent updates and contrastive discrimination.

\section{\bench}
\label{sec:bench}

This section introduces \bench, a benchmark for evaluating action-conditioned multimodal time series forecasting.

\subsection{Task Formulation}
\label{sec:task}

Given $L$ historical observations of a univariate time series $\mathbf{x}=(x_{t-L+1},\ldots,x_t)\in\mathbb{R}^{L}$, textual context $c$ describing the system, and a textual action $a$ specifying an intervention and its temporal scope, the task is to learn a world model $\mathcal{F}$ that predicts the next $H$ values $\mathbf{y}=(x_{t+1},\ldots,x_{t+H})\in\mathbb{R}^{H}$ under the specified action, formulated as $\widehat{\mathbf{y}}=\mathcal{F}(\mathbf{x},c,a)$.

\subsection{Datasets}
\label{sec:datasets}

\bench\xspace comprises seven real-world datasets spanning three significant domains: economics, physical systems, and healthcare.
In \emph{economics}, M5~\cite{m5data} pairs daily item-level retail sales with recorded changes in weekly average selling prices.
MOF~\cite{mofdata} links Japan's foreign-exchange interventions, specifying traded currencies, transaction direction, and amount, to subsequent yen-per-dollar exchange rates.
FOMC~\cite{fomcdata} pairs U.S. monetary policy statements with subsequent two-year Treasury yields.
In \emph{physical systems}, Causal Chambers~\cite{causalchambersdata} records downstream air pressure in a wind tunnel under changes in inlet fan load.
DaISy~\cite{daisydata} pairs changes in liquid flow through a heat exchanger with its outlet temperature measurements.
In \emph{healthcare}, Shanghai Diabetes~\cite{shanghaidata} links continuous glucose measurements to clinical records of meals, non-insulin medication, and insulin administration.
For HUPA-UCM~\cite{hupadata}, we use raw FreeStyle glucose records at approximately 15-minute intervals, paired with recorded carbohydrate intake and insulin doses in people with type 1 diabetes.

We construct each instance by aligning recorded interventions with preceding time series histories and subsequent observations, consolidating simultaneous operations into a single action where applicable.
We then generate textual action descriptions from source records: FOMC retains its original policy statements, while the other datasets use standardized templates that preserve intervention details and timing.
System-level context is constructed separately using only information available at the forecast origin.
These components are organized into a unified sample format across datasets, providing a consistent interface for textual action conditioning.
Finally, dataset-specific preprocessing and validation ensure temporal consistency, filter incomplete action records and invalid observations, and produce well-aligned action--outcome samples.

\begin{table}[t]
\centering
\caption{Averaged forecasting performance across seven datasets. \textbf{Rank} is the mean rank over the three metrics. The best and second-best results in each column are shown in \textcolor{red}{\textbf{red}} and \textcolor{blue}{\textbf{blue}}, respectively.}
\label{tab:benchmark-overall}
\small
\begin{tabular}{c|ccc|c}
\toprule
\textbf{Model} & \textbf{nMAE} $\downarrow$ & \textbf{nRMSE} $\downarrow$ & \textbf{MASE} $\downarrow$ & \textbf{Rank} $\downarrow$ \\
\midrule
\textbf{TimesFM 3.0} & 1.0540 & 1.2087 & 3.5023 & 8.33 \\
\textbf{Chronos-2} & 1.0237 & 1.1695 & 3.4948 & 5.00 \\
\textbf{Toto 2.0} & 1.0257 & 1.1730 & 3.4923 & 5.33 \\
\textbf{Moirai 2.0} & 1.0329 & 1.1759 & 3.4996 & 7.00 \\
\midrule
\textbf{Time-LLM} & 1.4213 & 1.5774 & 4.6371 & 12.67 \\
\textbf{MM-TSFlib} & 1.1301 & 1.2530 & 3.8602 & 11.00 \\
\textbf{TimeCMA} & 1.4410 & 1.6115 & 5.3578 & 14.33 \\
\textbf{TADiff} & 1.4515 & 1.6465 & 5.0358 & 14.67 \\
\textbf{ChatTime} & 1.3114 & 1.4571 & 4.6900 & 12.33 \\
\textbf{Aurora} & 1.1108 & 1.2301 & 3.7131 & 10.00 \\
\midrule
\textbf{CovTSWM} & 1.0583 & 1.1930 & 3.6517 & 8.67 \\
\midrule[\heavyrulewidth]
\textbf{TSFM++$_{\,/\,\text{TimesFM 3.0}}$} & \textcolor{red}{\textbf{0.9983}} & \textcolor{red}{\textbf{1.1116}} & \textcolor{red}{\textbf{3.3634}} & \textcolor{red}{\textbf{1.00}} \\
\textbf{TSFM++$_{\,/\,\text{Chronos-2}}$} & 1.0140 & \textcolor{blue}{\textbf{1.1135}} & \textcolor{blue}{\textbf{3.4107}} & \textcolor{blue}{\textbf{2.33}} \\
\textbf{TSFM++$_{\,/\,\text{Toto 2.0}}$} & 1.0310 & 1.1603 & 3.4827 & 4.67 \\
\textbf{TSFM++$_{\,/\,\text{Moirai 2.0}}$} & \textcolor{blue}{\textbf{1.0121}} & 1.1298 & 3.4428 & 2.67 \\
\bottomrule
\end{tabular}
\end{table}

\subsection{Evaluation Protocol}
\label{sec:baselines}

\noindent\textbf{Baselines.}
We evaluate four classes of baselines, covering zero-shot forecasting and dataset-specific training:
(1) \emph{Time Series Foundation Models (TSFMs)}: TimesFM~3.0~\cite{timesfm32026}, Chronos-2~\cite{chronos22025}, Toto~2.0~\cite{toto22026}, and Moirai~2.0~\cite{moirai22025} provide zero-shot forecasts from numerical histories, establishing the performance of pretrained temporal models without action information.
(2) \emph{Multimodal Time Series Models}: Time-LLM~\cite{timellm2024}, MM-TSFlib from Time-MMD~\cite{timemmd2024}, TimeCMA~\cite{timecma2025}, and TADiff~\cite{tadiff2026} are trained on each dataset to evaluate task-specific numerical--textual integration for action-conditioned forecasting.
(3) \emph{Multimodal TSFMs}: ChatTime~\cite{chattime2025} and Aurora~\cite{aurora2026} are evaluated zero-shot to assess whether pretrained multimodal representations transfer to textual action conditioning.
(4) \emph{Covariate-aware TSWM}: CovTSWM (our name for the method in~\cite{zhang2026on} ) is trained on each dataset with an adapted covariate schema that encodes recorded actions as structured controls.
We use the best-performing module combinations for the modular frameworks: MM-TSFlib and CovTSWM, with implementation details in \appref{app:baseline-details}.
All text-capable baselines receive equivalent context and action information through their respective input interfaces.

\noindent\textbf{Metrics.}
We report normalized mean absolute error (nMAE) and normalized root mean squared error (nRMSE), scaled by the standard deviation of each input history, and mean absolute scaled error (MASE), scaled by its mean absolute lag-one difference, with lower values indicating better forecasts.
All scales use observed historical values only, and zero-scale instances are excluded rather than assigned zero error (\appref{app:metrics}).
Additional implementation details appear in \secref{sec:exp-setup}.

\begin{figure*}[!t]
    \centering
    \includegraphics[width=\textwidth]{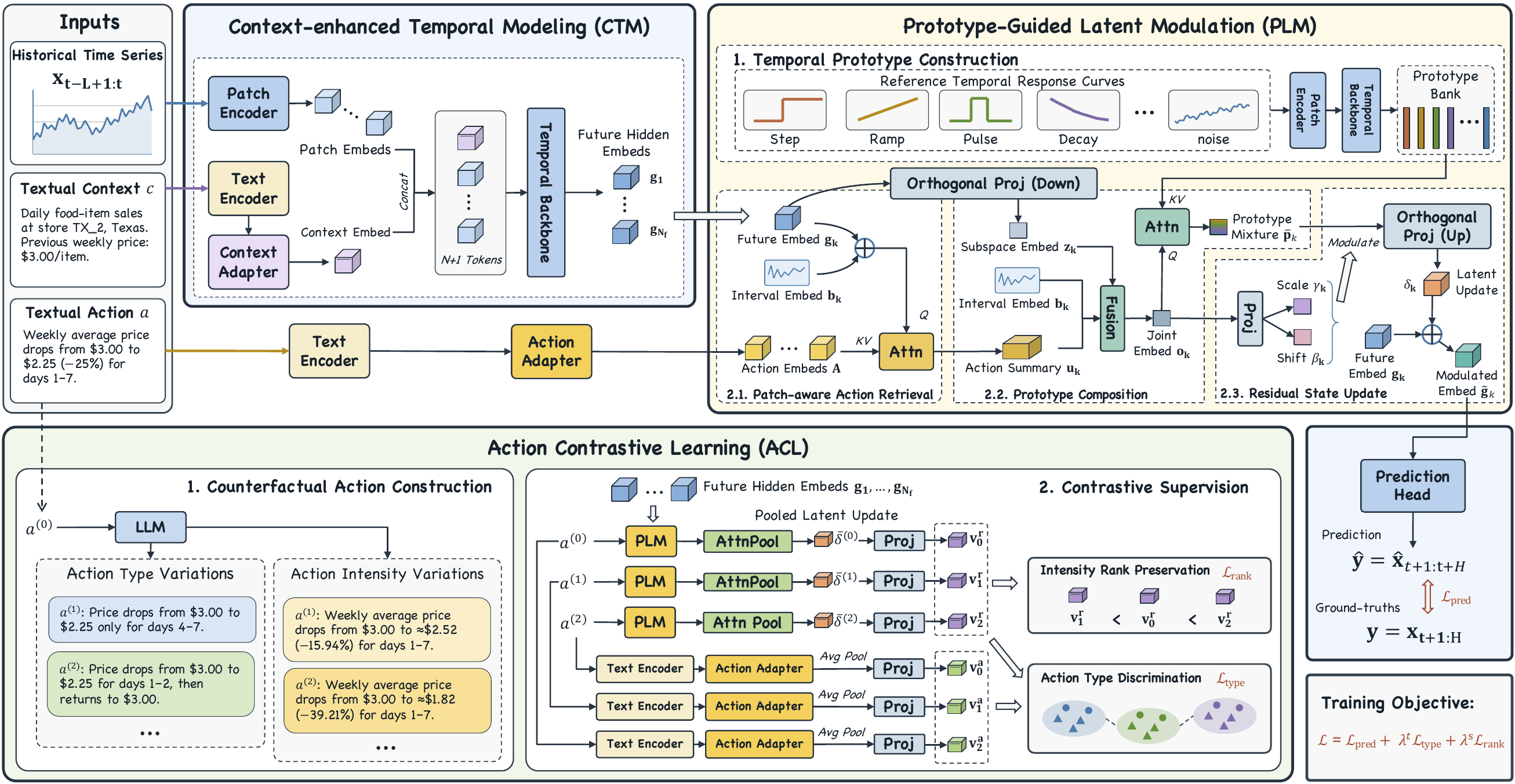}
    \caption{Overview of \model.}
    \label{fig:main}
\end{figure*}

\subsection{Benchmark Findings}
\label{sec:benchmark-findings}

\tabref{tab:benchmark-overall} shows that both dataset-trained multimodal models and zero-shot multimodal foundation models have higher aggregate errors than history-only TSFMs, suggesting that access to textual actions alone does not ensure effective forecasting.
CovTSWM achieves lower errors than these multimodal baselines across all three metrics, supporting the effectiveness of explicit action modeling.
Yet it remains less accurate than the strongest TSFMs, underscoring the predictive value of pretrained world-dynamics knowledge.
These findings raise a central question for world modeling: \emph{can pretrained TSFMs build on their implicit world dynamics to explicitly model how textual actions shape future trajectories}?

We thereby introduce TSFM++, a dataset-specific adaptation applicable to each evaluated TSFM that prepends projected context and action embeddings to its time series tokens (illustrated in \secref{sec:backbone}).
We denote backbone-specific variants by TSFM++$_{\,/\,\text{backbone}}$.
As shown in \tabref{tab:benchmark-overall}, action-prefix adaptation generally reduces aggregate errors, although the gains vary across backbones and Toto~2.0 does not improve in nMAE.
These results suggest that textual actions can refine pretrained predictive dynamics, motivating \model's explicit action-dependent latent updates, as detailed in \secref{sec:model}.
Additional results in \appref{app:covariate-conditioning} highlight the advantage of textual action conditioning, with TSFM++ outperforming TimesFM~3.0 using native structured covariates under both zero-shot and fine-tuning settings.

\section{\model}
\label{sec:model}

\figref{fig:main} illustrates the overall architecture of \model.
First, \emph{Context-enhanced Temporal Modeling (CTM)} incorporates textual context into a pretrained TSFM to obtain context-aware temporal representations.
Next, \emph{Prototype-Guided Latent Modulation (PLM)} learns a bank of temporal response prototypes and adaptively uses them to update the TSFM hidden states according to the specified action.
The original TSFM prediction head then decodes these action-adjusted states into future values.
To complement the forecasting loss, \emph{Action Contrastive Learning (ACL)} further guides the latent updates to distinguish action types and preserve the ordering of action intensities.

\subsection{Context-enhanced Temporal Modeling}
\label{sec:backbone}
To leverage implicit world knowledge of temporal dynamics, we build \model\xspace upon a pretrained TSFM consisting of three main components: a patch encoder $\operatorname{Enc}$, a temporal backbone $\operatorname{Backbone}$, and a prediction head $\operatorname{Dec}$.

Given a historical time series $\mathbf{x}$ of length $L$, the patch encoder divides it into patches of length $L_{\mathrm{in}}$ and maps them into embeddings $\mathbf{E}=\operatorname{Enc}(\mathbf{x})\in\mathbb{R}^{N\times D}$, where $N=\lceil L/L_{\mathrm{in}}\rceil$ is the number of input patches and $D$ is the hidden dimension.
To capture domain-specific dynamics that may not be identifiable from numerical observations alone, we additionally introduce textual context $c$.
Specifically, a pretrained text embedding model $\operatorname{TextEnc}^c$ encodes $c$ into a vector, which is projected into the TSFM hidden space by a lightweight adapter $\operatorname{Adapt}^c$ comprising two linear layers and a GELU activation:
$\mathbf{c}=\operatorname{Adapt}^c(\operatorname{TextEnc}^c(c))\in\mathbb{R}^{D}$.
We prepend this context embedding as an additional token to the numerical patch sequence, obtaining $\mathbf{E}^c=[\mathbf{c};\mathbf{E}]\in\mathbb{R}^{(N+1)\times D}$, which is processed by the temporal backbone using its native forecasting mechanism.

For a forecasting horizon of $H$ time steps, the TSFM predicts $N_{\mathrm{f}}=\lceil H/L_{\mathrm{out}}\rceil$ future patches of length $L_{\mathrm{out}}$, each associated with a hidden state $\mathbf{g}_k\in\mathbb{R}^{D}$, where $k=1,\ldots,N_{\mathrm{f}}$ indexes future patches.
Different TSFM architectures obtain these states through distinct forecasting paradigms.
Autoregressive models generate future patches sequentially, using previously predicted patches to compute the next hidden state: $\mathbf{g}_k = \operatorname{Backbone}([\mathbf{E}^c;\operatorname{Enc}(\widehat{\mathbf{y}}_{<k})])_{\mathrm{last}}$,
where $\widehat{\mathbf{y}}_{<k}$ denotes the preceding predicted patches and the subscript $\mathrm{last}$ selects the final hidden state.
In contrast, masked-future models obtain all future hidden states jointly by appending $N_{\mathrm{f}}$ masked tokens $\mathbf{E}^{\mathrm{mask}}\in\mathbb{R}^{N_{\mathrm{f}}\times D}$, which is formulated as $[\mathbf{g}_1;\ldots;\mathbf{g}_{N_{\mathrm{f}}}] = \operatorname{Backbone}([\mathbf{E}^c;\mathbf{E}^{\mathrm{mask}}])_{\mathrm{future}}$,
where $\mathrm{future}$ selects the hidden states at the masked future positions.
In either paradigm, the TSFM yields context-aware but action-unaware future states $\mathbf{g}_k$, which our subsequent \emph{PLM} module updates into action-conditioned states $\widetilde{\mathbf{g}}_k$.
They are finally decoded by the prediction head as $\widehat{\mathbf{y}}_k=\operatorname{Dec}(\widetilde{\mathbf{g}}_k)$.

For comparison, TSFM++, the action-prefix baseline in \secref{sec:benchmark-findings}, directly prepends adapted embeddings of both context and action to the numerical patch sequence, without \emph{PLM} or \emph{ACL} modules.

\subsection{Prototype-Guided Latent Modulation}
\label{sec:modulation}

To ground textual interventions in pretrained world dynamics, \emph{PLM} composes temporal response prototypes as building blocks to modulate the TSFM hidden states.
It consists of two components: \emph{(1) Temporal Prototype Construction}, which builds a bank of representative temporal patterns, and \emph{(2) Patch-wise Modulation}, which adaptively combines these prototypes to produce action-dependent latent updates for future forecasts.

\subsubsection{Temporal Prototype Construction}
\label{sec:prototypes}

To capture diverse intervention effects, we construct $N_{\mathrm{p}}$ reference curves representing typical temporal responses (\eg steps, ramps, pulses) with variations in onset, duration, and rate.
Each curve is normalized and resampled to the input patch length $L_{\mathrm{in}}$, yielding $\mathbf{x}^{\mathrm{ref}}_m\in\mathbb{R}^{L_{\mathrm{in}}}$.
We encode each curve as a single patch using TSFM's frozen patch encoder and temporal backbone, followed by a trainable two-layer projection:
\begin{equation}
    \mathbf{p}_m=\operatorname{Proj}\!\left(
    \operatorname{Backbone}(\operatorname{Enc}(\mathbf{x}^{\mathrm{ref}}_m))\right)
    \in\mathbb{R}^{R},\qquad m=1,\ldots,N_{\mathrm{p}},
    \label{eq:prototypes}
\end{equation}
where $\operatorname{Backbone}$ returns the single-patch hidden state, and $R<D$ is the dimension of the latent update subspace introduced below.
The resulting prototype bank $\mathbf{P}=[\mathbf{p}_1;\ldots;\mathbf{p}_{N_{\mathrm{p}}}]\in\mathbb{R}^{N_{\mathrm{p}}\times R}$ provides reusable temporal prototype representations for action-conditioned future modulation.

\subsubsection{Patch-wise Modulation}
\label{sec:readout-control}
To preserve pretrained knowledge of temporal evolution while accommodating action-induced deviations, we constrain latent updates to an $R$-dimensional learnable subspace.
This subspace is spanned by an orthonormal basis matrix $\mathbf{Q}\in\mathbb{R}^{D\times R}$ satisfying $\mathbf{Q}^{\top}\mathbf{Q}=\operatorname{Id}_R$, where $\operatorname{Id}_R$ is the $R$-dimensional identity matrix. The basis is obtained through differentiable reduced QR decomposition of a trainable matrix $\boldsymbol{\Theta}\in\mathbb{R}^{D\times R}$.
For each future patch, modulation involves three steps: retrieving relevant action information, composing prototypes based on the action and temporal state, and applying a residual update.

\noindent\textbf{Patch-aware Action Retrieval.}
An action description may specify operations, magnitudes, and timing that affect different forecast intervals.
To selectively retrieve such information, we retain the token-level representations $\operatorname{TextEnc}^a(a)$ from the pretrained text encoder, rather than the pooled context representation used in \emph{CTM}.
A two-layer adapter $\operatorname{Adapt}^a$ produces TSFM-aligned action embeddings $\mathbf{A}=\operatorname{Adapt}^a(\operatorname{TextEnc}^a(a))\in\mathbb{R}^{N_{\mathrm{a}}\times D}$, where $N_{\mathrm{a}}$ is the number of action tokens.

For future patch $k$, we encode its start and end steps $(\mathrm{start}_k,\mathrm{end}_k)$ relative to the forecast origin using sinusoidal features and a linear projection, obtaining an interval embedding $\mathbf{b}_k\in\mathbb{R}^{D}$.
Combining this embedding with the normalized temporal state $\bar{\mathbf{g}}_k=\operatorname{LayerNorm}(\mathbf{g}_k)$, we retrieve relevant action information through cross-attention:
\begin{equation}
    \mathbf{u}_k=\operatorname{Attn}\!\left(
    \operatorname{Linear}(\bar{\mathbf{g}}_k)+\mathbf{b}_k,
    \mathbf{A}\right)\in\mathbb{R}^{D},
    \label{eq:action-retrieval}
\end{equation}
where $\operatorname{Linear}$ is a trainable $D$-dimensional projection, and $\operatorname{Attn}$ denotes cross-attention with trainable query, key, and value projections.
The resulting $\mathbf{u}_k$ summarizes action information relevant to the prediction interval of patch $k$.

\noindent\textbf{Prototype Composition.}
Since the response to an action also depends on the current temporal dynamics, we combine the retrieved action information with the hidden state to select relevant prototypes.
We first project $\mathbf{g}_k$ onto the learned subspace, obtaining its coordinates $\mathbf{z}_k=\mathbf{Q}^{\top}\mathbf{g}_k\in\mathbb{R}^{R}$ in the orthonormal basis $\mathbf{Q}$.
A two-layer MLP then integrates the subspace coordinates, action information, and interval embedding:
$\mathbf{o}_k=\operatorname{MLP}([\mathbf{z}_k;\mathbf{u}_k;\mathbf{b}_k])\in\mathbb{R}^{D}$.
Using this joint representation as the query, we retrieve a weighted mixture of temporal prototypes:
\begin{equation}
    \bar{\mathbf{p}}_k=\operatorname{Attn}\!\left(
    \mathbf{o}_k,\mathbf{P}\right)\in\mathbb{R}^{R},
    \label{eq:prototype-retrieval}
\end{equation}
where a separate cross-attention layer adaptively combines the prototypes according to the action and temporal state.

\noindent\textbf{Residual State Update.}
To flexibly adjust the selected response patterns, two linear heads map $\mathbf{o}_k$ to a signed scale $\boldsymbol{\gamma}_k\in\mathbb{R}^{R}$ and an additive shift $\boldsymbol{\beta}_k\in\mathbb{R}^{R}$.
The resulting update is mapped back to the TSFM hidden space and added to the original state:
\begin{equation}
    \boldsymbol{\delta}_k=\mathbf{Q}(\boldsymbol{\gamma}_k\odot\bar{\mathbf{p}}_k+\boldsymbol{\beta}_k),
    \qquad
    \widetilde{\mathbf{g}}_k=\mathbf{g}_k+\boldsymbol{\delta}_k,
    \label{eq:modulation}
\end{equation}
where $\odot$ denotes element-wise multiplication.
The signed scale controls the strength and direction of prototype contributions, while the shift provides additional state-dependent adjustments.
Since $\boldsymbol{\delta}_k$ lies in the subspace spanned by $\mathbf{Q}$, the update preserves the orthogonal component of the original hidden state.
The adjusted states $\{\widetilde{\mathbf{g}}_k\}_{k=1}^{N_{\mathrm{f}}}$ are decoded by the original prediction head $\operatorname{Dec}$, giving final predictions $\{\widehat{\mathbf{y}}_k\}_{k=1}^{N_{\mathrm{f}}}$.

\subsection{Action Contrastive Learning}
\label{sec:action-learning}

While \emph{PLM} achieves fine-grained action grounding, learning from factual actions alone may still be dominated by history patterns and overlook the specific characteristics of the given action.
To strengthen action sensitivity, \emph{ACL} explicitly contrasts the latent updates induced by alternative actions under the same history and context.
It consists of two components: \emph{(1) Counterfactual Action Construction}, which generates variations in action type and intensity, and \emph{(2) Contrastive Supervision}, which distinguishes action-specific responses and preserves their relative intensity relationships.

\subsubsection{Counterfactual Action Construction}
\label{sec:action-variation}

We construct two groups of alternative actions that vary in type or intensity, both including the factual action. To distinguish these candidates, we index the actions and write the original factual action $a$ as $a^{(0)}$.
For type variations, an LLM first summarizes the action categories in each dataset's training split according to their operations or temporal policies.
Given a factual action, it then selects $N_{\mathrm{t}}$ distinct categories other than the factual category and jointly generates one alternative per category, forming $\{a^{(i)}\}_{i=0}^{N_{\mathrm{t}}}$.
For intensity variations, we randomly assign the factual action a rank $\rho^{(0)}\in\{0,\ldots,N_{\mathrm{s}}\}$ among $N_{\mathrm{s}}+1$ levels ordered from weakest to strongest, and sample a percentage change between adjacent levels.
Given this rank and percentage, the LLM jointly generates $N_{\mathrm{s}}$ alternatives by progressively decreasing intervention magnitudes below the factual rank and increasing them above it, while retaining the action type.
Together with the unchanged factual action, these alternatives form $\{a^{(i)}\}_{i=0}^{N_{\mathrm{s}}}$.
In both cases, generation preserves the factual description's format and all content unrelated to the intended variation.
Further construction and validation details are provided in \appref{app:action-construction}.

\subsubsection{Contrastive Supervision}
\label{sec:increment-contrast}

Since alternative actions share the same historical observations and context, their hidden states may be dominated by common temporal information rather than action-specific effects.
We therefore apply contrastive supervision directly to the latent updates from \emph{PLM}, defined as $\boldsymbol{\delta}_k^{(i)}=\widetilde{\mathbf{g}}_k^{(i)}-\mathbf{g}_k^{(i)}$ for the $i$-th action and $k$-th future patch.

\noindent\textbf{Latent Response Aggregation.}
To capture each action's overall effect across the forecast horizon, we aggregate its patch-level latent updates through attention-weighted pooling:
\begin{equation}
    \bar{\boldsymbol{\delta}}^{(i)}=\sum_{k=1}^{N_{\mathrm{f}}}\omega_k^{(i)}\boldsymbol{\delta}_k^{(i)},
    \quad
    \omega_k^{(i)}=\operatorname{softmax}_{k}
    (\mathbf{w}^{\top}\boldsymbol{\delta}_k^{(i)}+\log\kappa_k),
    \label{eq:response-pooling}
\end{equation}
where $\mathbf{w}\in\mathbb{R}^{D}$ is trainable, $\kappa_k$ denotes the valid fraction of patch $k$ excluding padding, and $\operatorname{softmax}_k$ normalizes weights across future patches.
We also obtain an action representation $\bar{\mathbf{a}}^{(i)}$ by averaging the valid token embeddings in $\mathbf{A}^{(i)}$.
The pooled action and latent response representations are projected into a shared embedding space and $\ell_2$-normalized:
\begin{equation}
    \mathbf{v}_i^{\mathrm{a}}=\operatorname{Norm}(\operatorname{Proj}_{\mathrm{a}}(\bar{\mathbf{a}}^{(i)})),
    \qquad
    \mathbf{v}_i^{\mathrm{r}}=\operatorname{Norm}(\operatorname{Proj}_{\mathrm{r}}(\bar{\boldsymbol{\delta}}^{(i)})),
    \label{eq:response-embedding}
\end{equation}
where $\operatorname{Proj}_{\mathrm{a}}$ and $\operatorname{Proj}_{\mathrm{r}}$ are trainable linear projections to the same dimension, and $\operatorname{Norm}$ denotes $\ell_2$ normalization.

\noindent\textbf{Action Type Discrimination.}
To distinguish the effects of different action types, we align each action embedding with its corresponding latent response while separating it from responses to alternative actions.
For the $N_{\mathrm{t}}+1$ actions contrasted by type, we define the action-response similarity as $\xi^{\mathrm{t}}_{i,i'}=(\mathbf{v}_i^{\mathrm{a}})^{\top}\mathbf{v}_{i'}^{\mathrm{r}}/\tau^{\mathrm{t}}$, where $i$ indexes the action, $i'$ indexes the response, and $\tau^{\mathrm{t}}>0$ is the temperature.
We adopt a bidirectional contrastive objective:
\begin{equation}
\mathcal{L}_{\mathrm{type}}=-\frac{1}{2(N_{\mathrm{t}}+1)}
\sum_{i=0}^{N_{\mathrm{t}}}
\log\frac{\exp(2\xi^{\mathrm{t}}_{i,i})}
{(\sum_{i'=0}^{N_{\mathrm{t}}}\exp(\xi^{\mathrm{t}}_{i,i'}))(\sum_{i'=0}^{N_{\mathrm{t}}}\exp(\xi^{\mathrm{t}}_{i',i}))}.
\label{eq:type-loss}
\end{equation}
The two terms enforce action-to-response and response-to-action matching, respectively, using other action types under the same observations as negatives.

\noindent\textbf{Intensity Rank Preservation.}
Unlike distinct action types, intensity variations have an inherent ordering that should be reflected in their latent responses.
Inspired by Rank-N-Contrast~\cite{zha2023rank}, we encourage responses with closer intensity ranks to exhibit greater similarity.
For the $N_{\mathrm{s}}+1$ actions ordered by intensity, let $\rho^{(i)}$ denote the intensity rank of $a^{(i)}$, and define the rank distance $\operatorname{dist}_{i,i'}=|\rho^{(i)}-\rho^{(i')}|$ and response similarity $\xi^{\mathrm{s}}_{i,i'}=-\|\mathbf{v}_i^{\mathrm{r}}-\mathbf{v}_{i'}^{\mathrm{r}}\|_2/\tau^{\mathrm{s}}$, where $\tau^{\mathrm{s}}>0$ is the temperature.
For each anchor $i$ and candidate $i'$, the comparison set $\mathcal{S}_{i,i'}=\{i''\in\{0,\ldots,N_{\mathrm{s}}\}\setminus\{i\}:\operatorname{dist}_{i,i''}\geq\operatorname{dist}_{i,i'}\}$ contains candidates with equal or greater rank distances.
The ranking objective is
\begin{equation}
    \mathcal{L}_{\mathrm{rank}}=-\frac{1}{(N_{\mathrm{s}}+1)N_{\mathrm{s}}}
    \sum_{i=0}^{N_{\mathrm{s}}}\sum_{\substack{i'=0\\i'\ne i}}^{N_{\mathrm{s}}}
    \log\frac{\exp(\xi^{\mathrm{s}}_{i,i'})}
    {\sum_{i''\in\mathcal{S}_{i,i'}}\exp(\xi^{\mathrm{s}}_{i,i''})}.
    \label{eq:intensity-loss}
\end{equation}
This preserves relative intensity relationships in the latent updates without assuming a fixed direction of state change.

\begin{table*}[t]
\centering
\caption{Comparison results of four TSFM backbones. \textbf{Vanilla} performs zero-shot forecasting by TSFMs, and \textbf{Overall} averages the 13 settings across seven datasets. The best and second-best results of each metric are shown in \textcolor{red}{red} and \textcolor{blue}{blue}, respectively. \underline{Underlining} marks the best result for each TSFM backbone.}
\label{tab:comparison-inverse}
\resizebox{\textwidth}{!}{%
\begin{tabular}{c|c|ccc|ccc|ccc|ccc}
\toprule
\multirow{2}{*}{\textbf{Dataset}} & \multirow{2}{*}{\textbf{Metric}} & \multicolumn{3}{c|}{\textbf{TimesFM 3.0}} & \multicolumn{3}{c|}{\textbf{Chronos-2}} & \multicolumn{3}{c|}{\textbf{Toto 2.0}} & \multicolumn{3}{c}{\textbf{Moirai 2.0}} \\
\cmidrule{3-5}\cmidrule{6-8}\cmidrule{9-11}\cmidrule{12-14}
 & & \textbf{Vanilla} & \textbf{TSFM++} & \textbf{\model} & \textbf{Vanilla} & \textbf{TSFM++} & \textbf{\model} & \textbf{Vanilla} & \textbf{TSFM++} & \textbf{\model} & \textbf{Vanilla} & \textbf{TSFM++} & \textbf{\model} \\
\midrule
\multirow{3}{*}{\textbf{M5}} & \textbf{nMAE} & 1.1612 & \underline{0.9658} & 0.9803 & \underline{0.8784} & 1.0204 & 0.9829 & \textcolor{blue}{\textbf{\underline{0.8705}}} & 0.9255 & 0.9917 & \textcolor{red}{\textbf{\underline{0.8602}}} & 0.9788 & 1.0219 \\
 & \textbf{nRMSE} & 1.7597 & 1.3084 & \underline{1.2997} & 1.4231 & \textcolor{red}{\textbf{\underline{1.2914}}} & \textcolor{blue}{\textbf{1.2991}} & 1.4190 & 1.3742 & \underline{1.3102} & 1.4071 & 1.3476 & \underline{1.3002} \\
 & \textbf{MASE} & 2.4538 & \underline{2.0766} & 2.1151 & \underline{1.9042} & 2.2255 & 2.1294 & \textcolor{blue}{\textbf{\underline{1.8886}}} & 1.9649 & 2.0875 & \textcolor{red}{\textbf{\underline{1.8606}}} & 2.1473 & 2.1837 \\
\midrule
\multirow{3}{*}{\textbf{MOF}} & \textbf{nMAE} & 0.7793 & \underline{0.7787} & 0.8153 & 0.8019 & 0.7569 & \textcolor{blue}{\textbf{\underline{0.7347}}} & 0.7673 & 0.7908 & \underline{0.7638} & \textcolor{red}{\textbf{\underline{0.7235}}} & 0.7414 & 0.7527 \\
 & \textbf{nRMSE} & 0.8066 & \underline{0.8012} & 0.8402 & 0.8286 & 0.7845 & \textcolor{blue}{\textbf{\underline{0.7604}}} & 0.7958 & 0.8178 & \underline{0.7923} & \textcolor{red}{\textbf{\underline{0.7488}}} & 0.7677 & 0.7772 \\
 & \textbf{MASE} & \underline{3.2255} & 3.2568 & 3.4272 & 3.3110 & 3.0944 & \textcolor{red}{\textbf{\underline{3.0290}}} & \underline{3.1208} & 3.1972 & 3.1621 & \textcolor{blue}{\textbf{\underline{3.0477}}} & 3.1069 & 3.1953 \\
\midrule
\multirow{3}{*}{\textbf{FOMC}} & \textbf{nMAE} & 0.9576 & 0.9530 & \underline{0.9458} & 0.9239 & 0.9177 & \underline{0.9143} & \textcolor{red}{\textbf{\underline{0.8857}}} & 0.9008 & \textcolor{blue}{\textbf{0.8986}} & 0.9293 & 0.9223 & \underline{0.9059} \\
 & \textbf{nRMSE} & 0.9810 & 0.9750 & \underline{0.9718} & 0.9480 & 0.9400 & \underline{0.9382} & \textcolor{red}{\textbf{\underline{0.9132}}} & 0.9266 & \textcolor{blue}{\textbf{0.9247}} & 0.9533 & 0.9455 & \underline{0.9316} \\
 & \textbf{MASE} & 2.5115 & 2.5069 & \underline{2.4799} & 2.3817 & 2.3699 & \underline{2.3408} & \textcolor{red}{\textbf{\underline{2.2863}}} & 2.3335 & 2.3314 & 2.3935 & 2.3728 & \textcolor{blue}{\textbf{\underline{2.3112}}} \\
\midrule
\multirow{3}{*}{\makecell[c]{\textbf{Causal}\\\textbf{Chambers}}} & \textbf{nMAE} & 4.4275 & 4.4563 & \textcolor{red}{\textbf{\underline{3.2116}}} & 4.5003 & 4.6659 & \underline{4.4761} & 4.7793 & 4.7962 & \textcolor{blue}{\textbf{\underline{4.0175}}} & 4.8220 & 4.5824 & \underline{4.1415} \\
 & \textbf{nRMSE} & 4.7647 & 4.8067 & \textcolor{red}{\textbf{\underline{3.6509}}} & 4.8322 & 4.9809 & \underline{4.7993} & 5.1062 & 5.1089 & \textcolor{blue}{\textbf{\underline{4.3950}}} & 5.1251 & 4.9205 & \underline{4.4478} \\
 & \textbf{MASE} & 13.2580 & 13.3632 & \textcolor{red}{\textbf{\underline{9.3159}}} & 13.8178 & 14.2030 & \underline{13.6584} & 14.6407 & 14.6188 & \textcolor{blue}{\textbf{\underline{12.1117}}} & 14.6040 & 14.1977 & \underline{12.4113} \\
\midrule
\multirow{3}{*}{\textbf{DaISy}} & \textbf{nMAE} & 0.4025 & 0.2971 & \textcolor{red}{\textbf{\underline{0.2289}}} & 0.4191 & 0.2843 & \underline{0.2437} & 0.3957 & 0.3536 & \textcolor{blue}{\textbf{\underline{0.2430}}} & 0.4212 & 0.3348 & \underline{0.2506} \\
 & \textbf{nRMSE} & 0.4025 & 0.2971 & \textcolor{red}{\textbf{\underline{0.2289}}} & 0.4191 & 0.2843 & \underline{0.2437} & 0.3957 & 0.3536 & \textcolor{blue}{\textbf{\underline{0.2430}}} & 0.4212 & 0.3348 & \underline{0.2506} \\
 & \textbf{MASE} & 0.9123 & 0.6732 & \textcolor{red}{\textbf{\underline{0.5183}}} & 0.9495 & 0.6439 & \underline{0.5522} & 0.8966 & 0.8018 & \textcolor{blue}{\textbf{\underline{0.5507}}} & 0.9544 & 0.7595 & \underline{0.5661} \\
\midrule
\multirow{3}{*}{\makecell[c]{\textbf{Shanghai}\\\textbf{Diabetes}}} & \textbf{nMAE} & 0.7221 & 0.6725 & \underline{0.6662} & 0.7368 & 0.6703 & \textcolor{red}{\textbf{\underline{0.6418}}} & 0.7383 & 0.7054 & \underline{0.6866} & 0.7212 & 0.6907 & \textcolor{blue}{\textbf{\underline{0.6563}}} \\
 & \textbf{nRMSE} & 0.7978 & 0.7435 & \underline{0.7360} & 0.8129 & 0.7405 & \textcolor{red}{\textbf{\underline{0.7091}}} & 0.8166 & 0.7774 & \underline{0.7565} & 0.7971 & 0.7624 & \textcolor{blue}{\textbf{\underline{0.7240}}} \\
 & \textbf{MASE} & 3.9093 & \underline{3.6565} & 3.6637 & 3.9866 & 3.6276 & \textcolor{red}{\textbf{\underline{3.4871}}} & 3.9903 & 3.8338 & \underline{3.7984} & 3.8950 & 3.7245 & \textcolor{blue}{\textbf{\underline{3.5913}}} \\
\midrule
\multirow{3}{*}{\textbf{HUPA-UCM}} & \textbf{nMAE} & 0.6148 & \textcolor{blue}{\textbf{\underline{0.5939}}} & 0.6057 & 0.6441 & 0.6083 & \underline{0.6023} & 0.6197 & 0.6276 & \textcolor{red}{\textbf{\underline{0.5923}}} & 0.6474 & 0.6193 & \underline{0.6048} \\
 & \textbf{nRMSE} & 0.7264 & \textcolor{red}{\textbf{\underline{0.6969}}} & 0.7181 & 0.7540 & 0.7069 & \underline{0.7022} & 0.7308 & 0.7380 & \textcolor{blue}{\textbf{\underline{0.6972}}} & 0.7536 & 0.7256 & \underline{0.7063} \\
 & \textbf{MASE} & 3.1235 & \textcolor{blue}{\textbf{\underline{3.0106}}} & 3.0610 & 3.2746 & 3.1067 & \underline{3.0698} & 3.1974 & 3.1969 & \textcolor{red}{\textbf{\underline{3.0060}}} & 3.2944 & 3.1684 & \underline{3.1117} \\
\midrule[\heavyrulewidth]
\multirow{3}{*}{\textbf{Overall}} & \textbf{nMAE} & 1.0540 & 0.9983 & \textcolor{red}{\textbf{\underline{0.8997}}} & 1.0237 & 1.0140 & \underline{0.9781} & 1.0257 & 1.0310 & \textcolor{blue}{\textbf{\underline{0.9515}}} & 1.0329 & 1.0121 & \underline{0.9635} \\
 & \textbf{nRMSE} & 1.2087 & 1.1116 & \textcolor{red}{\textbf{\underline{1.0185}}} & 1.1695 & 1.1135 & \underline{1.0850} & 1.1730 & 1.1603 & \underline{1.0648} & 1.1759 & 1.1298 & \textcolor{blue}{\textbf{\underline{1.0636}}} \\
 & \textbf{MASE} & 3.5023 & 3.3634 & \textcolor{red}{\textbf{\underline{3.0651}}} & 3.4948 & 3.4107 & \underline{3.2981} & 3.4923 & 3.4827 & \textcolor{blue}{\textbf{\underline{3.2295}}} & 3.4996 & 3.4428 & \underline{3.2562} \\
\bottomrule
\end{tabular}%
}
\end{table*}

\subsection{Model Optimization}
\label{sec:optimization}

For a single instance, we jointly optimize factual forecasting accuracy and the two counterfactual contrastive objectives:
\begin{equation}
    \mathcal{L}=\mathcal{L}_{\mathrm{pred}}
    +\lambda^{\mathrm{t}}\mathcal{L}_{\mathrm{type}}
    +\lambda^{\mathrm{s}}\mathcal{L}_{\mathrm{rank}},
    \label{eq:total-loss}
\end{equation}
where $\lambda^{\mathrm{t}},\lambda^{\mathrm{s}}\geq0$ balance the contrastive objectives; the full training objective averages $\mathcal{L}$ over all training instances.
The forecasting loss is $\mathcal{L}_{\mathrm{pred}}=\frac{1}{H\sigma_{\mathbf{x}}^2}\sum_{j=1}^{H}(\widehat{y}_{j}-y_{j})^2$, where $j$ indexes forecast steps and $\sigma_{\mathbf{x}}$ is the historical standard deviation.
The pretrained TSFM and text encoders $\operatorname{TextEnc}^c,\operatorname{TextEnc}^a$ remain frozen, while the adapters, \emph{PLM}, and contrastive projections are jointly optimized.
\appref{app:backbone-adaptation} further examines whether fine-tuning the TSFM yields additional gains.
Counterfactual branches and contrastive projections are used only during training, adding no inference overhead.
The computational complexity is detailed in \appref{app:complexity}.

\section{Experiments}

\subsection{Experimental Setup}
\label{sec:exp-setup}

\noindent\textbf{Datasets.}
We use fixed chronological train/validation/test splits targeting a 60\%/ 20\%/20\% ratio.
History lengths $L$ and forecast horizons $H$ reflect each dataset's sampling frequency and intervention timescale.
Except for Causal Chambers, each dataset has two $(L,H)$ settings, whose performance is averaged.
Detailed descriptions of datasets are provided in \appref{app:datasets}.

\noindent\textbf{Hyperparameters.}
We retain each TSFM's native input and output patch lengths $L_{\mathrm{in}}$, $L_{\mathrm{out}}$ and hidden dimension $D$.
We set the modulation subspace dimension to $R=32$ and use $N_{\mathrm{p}}=24$ temporal prototypes, with $N_{\mathrm{t}}=N_{\mathrm{s}}=5$ alternatives for action type and intensity, respectively.
The type and intensity contrastive objectives use temperatures $\tau^{\mathrm{t}}=0.1$ and $\tau^{\mathrm{s}}=0.2$ and loss weights $\lambda^{\mathrm{t}}=\lambda^{\mathrm{s}}=0.01$.
The text adapters have a hidden width of 256, while the contrastive projections map to a shared 128-dimensional space.

\noindent\textbf{Training Configurations.}
We encode text with Qwen3-Embedding-4B~\cite{qwen3embedding} and generate counterfactual actions with GPT-6 Luna~\cite{gpt6luna2026}.
TSFM++ and \model\xspace are trained per dataset setting with frozen TSFMs and text encoders, using AdamW with a default learning rate of $10^{-4}$, weight decay of 0.01, and an effective batch size of 16.
We train for up to ten epochs, with early stopping after three epochs without validation improvement, and retain the checkpoint with the lowest validation forecasting loss.
All runs use NVIDIA H200 GPUs and a fixed seed of 42 to control training randomness and facilitate reproducibility.

\begin{figure}[h]
    \centering
    \includegraphics[width=\linewidth]{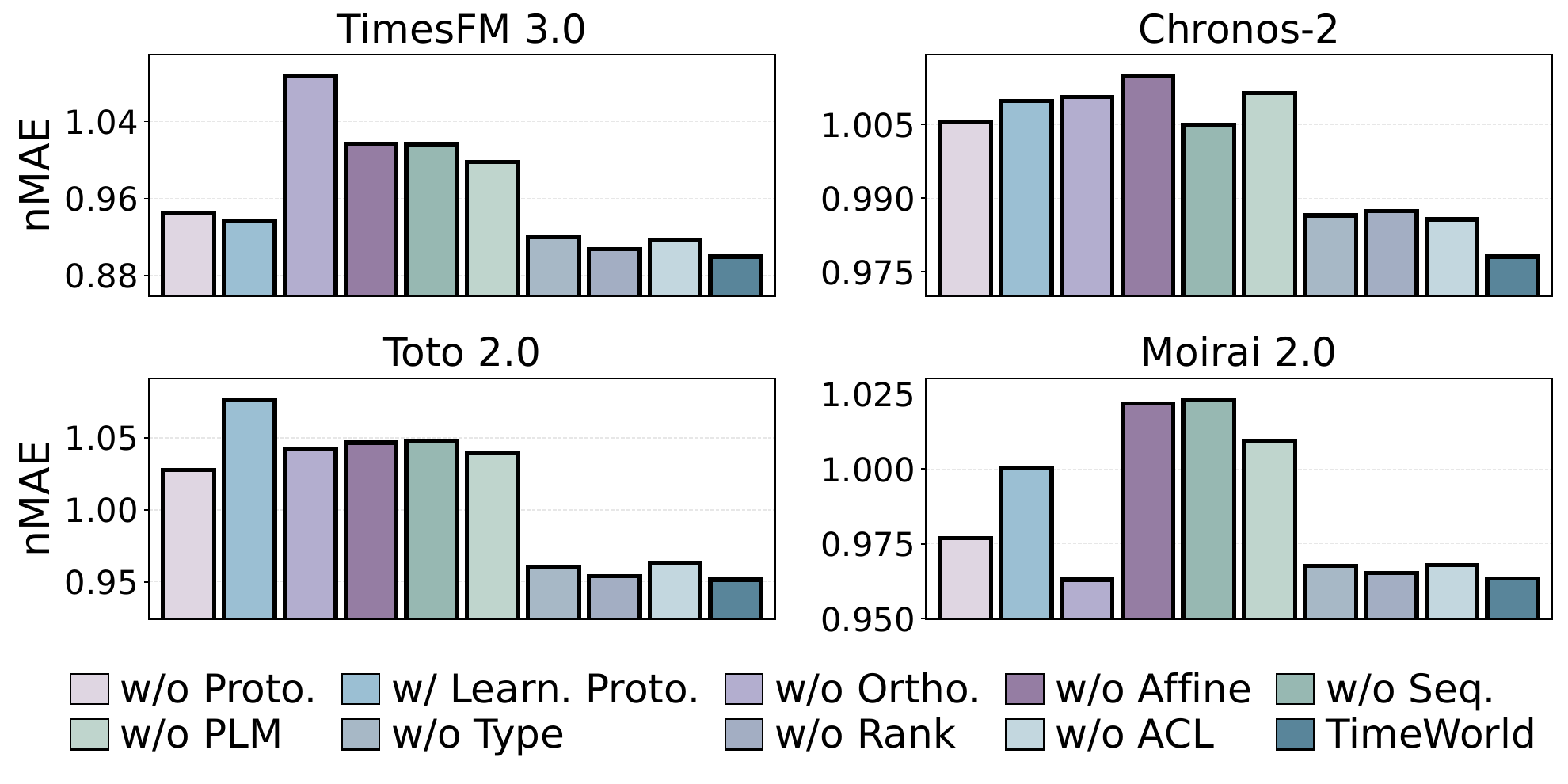}
    \caption{Ablation results of \model.}
    \label{fig:ablation}
\end{figure}

\begin{figure*}[t]
    \centering
    \subfloat[Type weight $\lambda^{\mathrm{t}}$.\label{fig:parameter-type}]{%
        \includegraphics[width=0.33\textwidth]{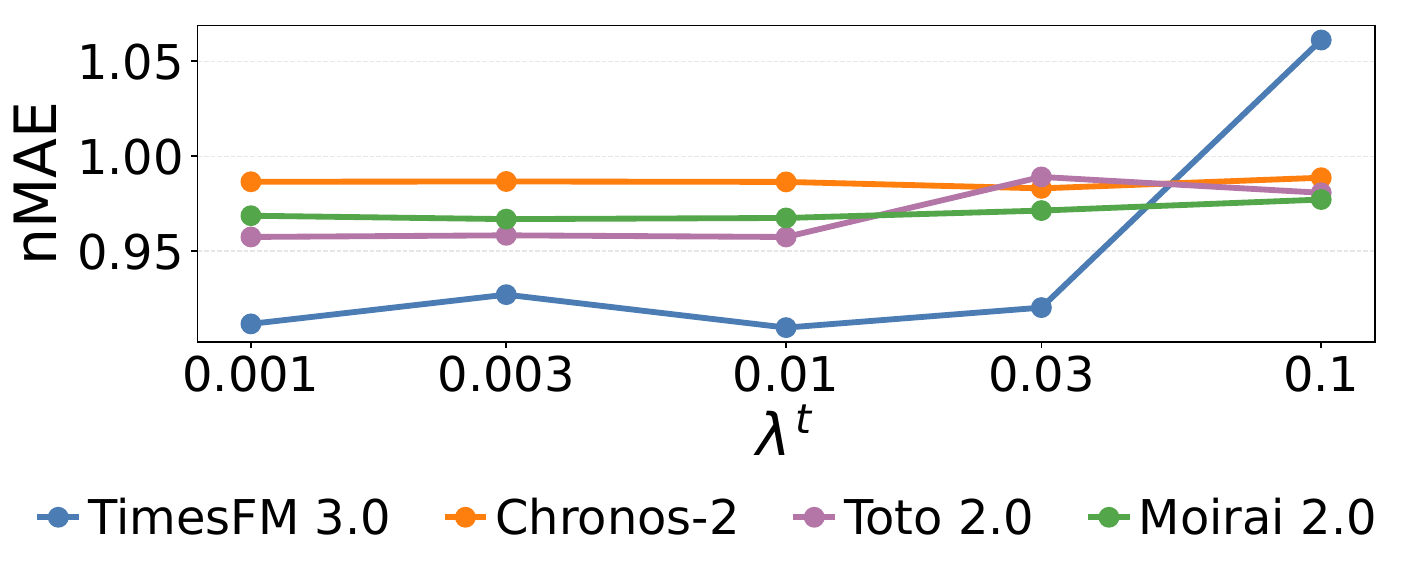}}\hfill
    \subfloat[Intensity weight $\lambda^{\mathrm{s}}$.\label{fig:parameter-intensity}]{%
        \includegraphics[width=0.33\textwidth]{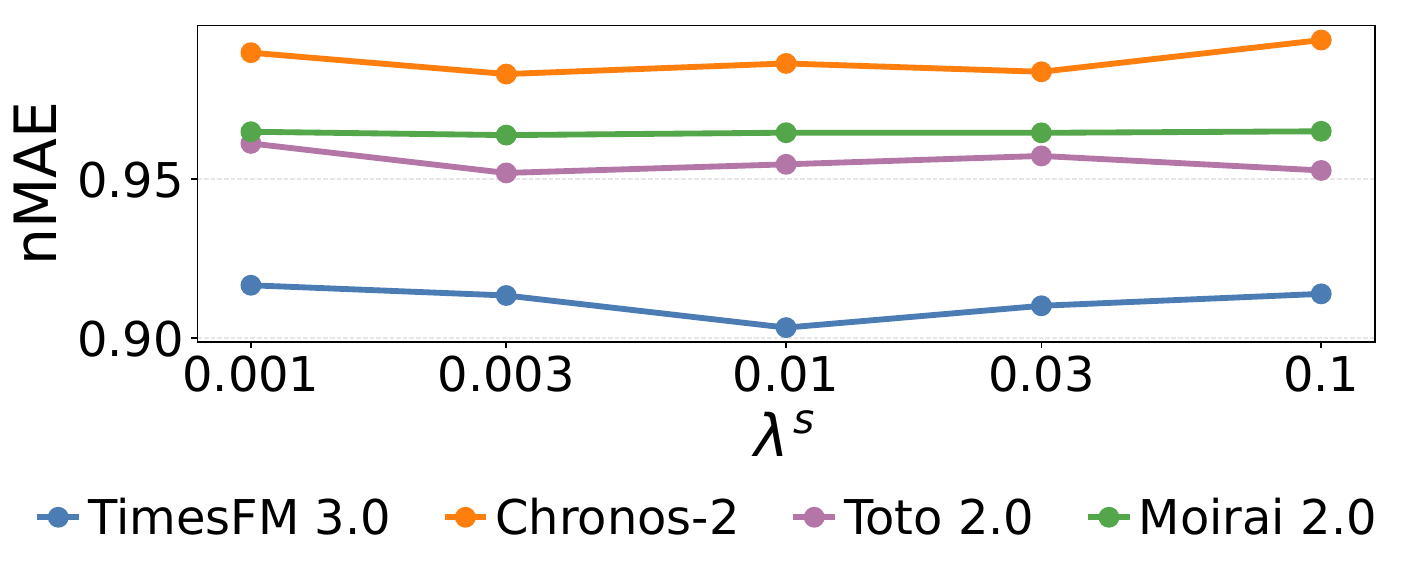}}\hfill
    \subfloat[Subspace dimension $R$.\label{fig:parameter-rank}]{%
        \includegraphics[width=0.33\textwidth]{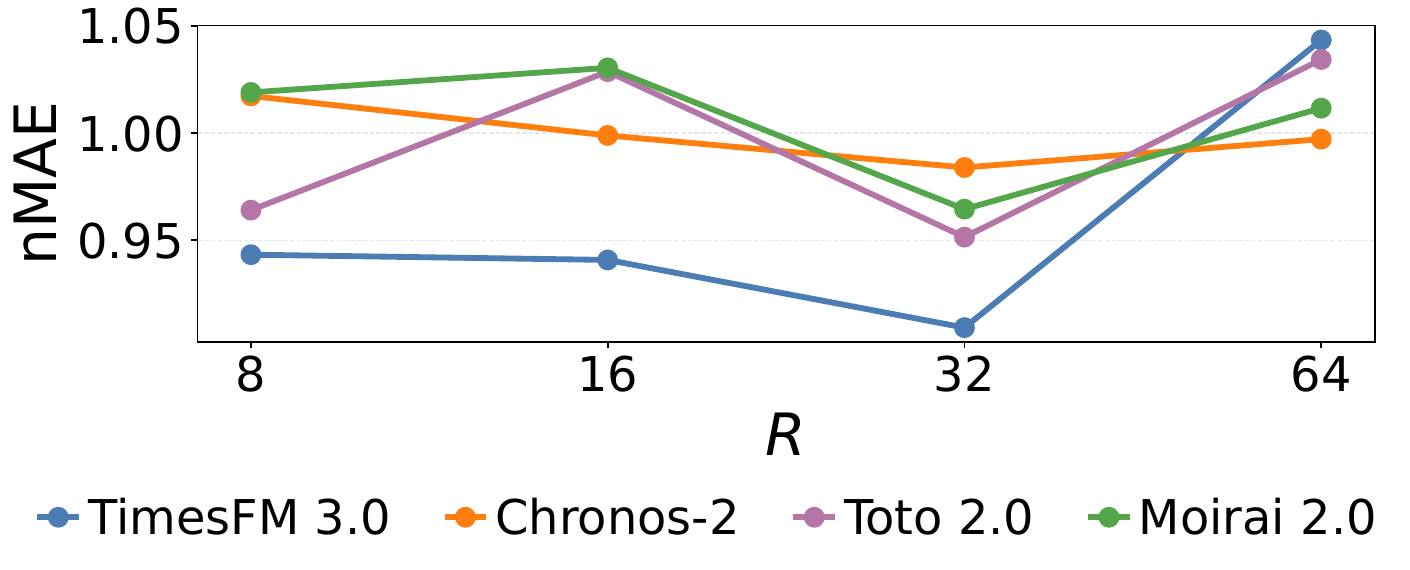}}
    \caption{Parameter sensitivity of \model\xspace across four TSFM backbones, measured by nMAE.}
    \label{fig:parameter}
\end{figure*}

\subsection{Comparison Results}
\label{sec:comparison}

\tabref{tab:comparison-inverse} compares \model\xspace with vanilla TSFMs and TSFM++, revealing three main observations:
(1) \emph{\model\xspace consistently improves aggregate forecasting performance across backbones.}
It reduces all three overall errors relative to both alternatives, with nMAE improvements of 3.5\%--9.9\% over TSFM++.
These gains support structured action-dependent updates to pretrained world dynamics beyond direct text-prefix conditioning.
(2) \emph{Stronger history-only forecasting does not imply greater benefits from action conditioning.}
TimesFM~3.0 has the highest overall errors among vanilla backbones but achieves the lowest errors with both TSFM++ and \model.
Its larger gains, compared with the more modest improvements of Chronos-2, indicate that the ability to exploit textual actions varies across pretrained backbones.
(3) \emph{The benefits of action modeling vary across domains.}
\model\xspace improves all three metrics over both alternatives on Causal Chambers and DaISy for every backbone.
Healthcare forecasting also generally improves, though TSFM++ remains stronger with TimesFM on HUPA-UCM.
Economic results are less consistent: improvements on FOMC are small, MOF results depend on the backbone, and M5 often favors vanilla forecasting in nMAE and MASE despite reductions in nRMSE.
Confidence intervals for the gains over vanilla TSFM forecasting are reported in \appref{app:forecast-confidence}.

\subsection{Ablation Study}
\label{sec:ablation}

In this section, we evaluate nine variants to isolate the contributions of \emph{PLM} and \emph{ACL}.
Within PLM, ``w/o Proto.'' directly modulates the subspace coordinates without prototypes, ``w/ Learn. Proto.'' replaces encoded prototypes with randomly initialized learnable vectors, ``w/o Ortho.'' replaces the orthogonal subspace projections with MLP mappings, ``w/o Affine'' removes the final scale and shift, and ``w/o Seq.'' replaces token-level action attention with a single action embedding.
``w/o PLM'' uses action-prefix conditioning instead of latent modulation, while ``w/o Type'', ``w/o Rank'', and ``w/o ACL'' remove type discrimination, intensity ranking, and both contrastive objectives, respectively.
\figref{fig:ablation} yields three observations:
(1) \emph{Structured latent modulation improves action grounding.}
Removing prototypes, affine adjustment, or token-level retrieval increases nMAE for every backbone, as does replacing PLM with action-prefix conditioning.
Learnable prototypes also worsen all four backbones, supporting the value of temporal response priors encoded by pretrained TSFMs.
These results support composing temporal response patterns and adapting them to the action and current state.
(2) \emph{The subspace parameterization matters, but its effect depends on the backbone.}
Switching it degrades TimesFM~3.0, Chronos-2, and Toto~2.0, while Moirai~2.0 changes only marginally in the opposite direction.
(3) \emph{Both contrastive objectives contribute to forecasting accuracy.}
Removing either objective or ACL as a whole worsens all four backbones, supporting complementary supervision of action type and intensity.

\subsection{Parameter Analysis}
\label{sec:parameter}

In this section, we examine the contrastive weights $\lambda^{\mathrm{t}}$ and $\lambda^{\mathrm{s}}$, which balance action discrimination against forecasting accuracy, and the subspace dimension $R$, which controls the capacity of latent modulation.
Each contrastive weight is varied independently over $\{0.001,0.003,0.01,0.03,0.1\}$, and $R$ over $\{8,16,32,64\}$, keeping the remaining settings fixed.
\figref{fig:parameter} shows that:
(1) \emph{Moderate type weights provide stable performance, but excessive weighting can be harmful.}
TimesFM is relatively stable up to $\lambda^{\mathrm{t}}=0.03$, whereas increasing it to 0.1 raises nMAE by 16.7\% relative to 0.01.
This may reflect overly strong pressure to separate action types at the expense of factual prediction.
(2) \emph{Performance is particularly robust to the intensity weight.}
Across the full range of $\lambda^{\mathrm{s}}$, each backbone's nMAE remains within 1.5\% of its best value, indicating limited sensitivity to the strength of rank supervision.
(3) \emph{An intermediate subspace dimension balances capacity and constraint.}
All four backbones achieve their lowest nMAE at $R=32$, supporting our chosen dimension.
Smaller subspaces may restrict useful action-dependent updates, while larger ones permit unnecessary changes to pretrained representations.

\subsection{Action Consistency Analysis}
\label{sec:action-consistency}

Forecast errors measure agreement with observed outcomes but do not reveal whether a world model distinguishes alternative actions.
We therefore examine representation similarity as a proxy for counterfactual action sensitivity, comparing \model\xspace with w/o ACL under the same history and context.
In \tabref{tab:action-consistency}, \emph{Same} and \emph{Different} measure cosine similarity between action embeddings and responses to the corresponding or alternative action types.
\emph{Adjacent} and \emph{Distant} compare response embeddings at neighboring and maximally separated intensity ranks.
Both models use a shared readout, and scores are averaged across backbones and datasets.
We observe that ACL increases the separation between matching and mismatched action types while making responses to nearby intensity ranks more similar than those to distant ranks. 
This supports more distinguishable counterfactual responses beyond improvements in factual forecasting.

\begin{table}[h]
\centering
\caption{Action-conditioned representation similarity.}
\label{tab:action-consistency}
\small
\setlength{\tabcolsep}{4pt}
\begin{tabular}{l|cc|cc}
\toprule
\multirow{2}{*}{\textbf{Model}} & \multicolumn{2}{c|}{\textbf{Action Type}} & \multicolumn{2}{c}{\textbf{Action Intensity}} \\
\cmidrule(lr){2-3}\cmidrule(lr){4-5}
 & \textbf{Same} & \textbf{Different} & \textbf{Adjacent} & \textbf{Distant} \\
\midrule
\textbf{\model} & 0.1077 & 0.0533 & 0.9815 & 0.9380 \\
\textbf{w/o ACL} & 0.0293 & 0.0271 & 0.9311 & 0.9284 \\
\bottomrule
\end{tabular}
\end{table}

We further examine the effectiveness of prototype composition in \appref{app:prototype-composition} and provide a qualitative case study in \appref{app:case-study}.

\section{Conclusion}
\label{sec:conclusion}

In this work, we study time series world models that specify interventions through textual actions beyond fixed covariate schemas, and introduce \bench\xspace for evaluation across seven real-world datasets.
Our benchmark findings suggest that pretrained TSFMs encode implicit world knowledge of temporal evolution, providing a foundation for action-conditioned world modeling. 
Building on this insight, we propose \model, a plug-in framework that adapts pretrained TSFMs through explicit action grounding and discrimination.
Prototype-guided latent modulation translates textual actions into state- and interval-dependent updates, while action contrastive learning distinguishes responses to different action types and preserves relative intensity relationships under shared histories.
Experiments across four TSFM backbones demonstrate consistent improvements in aggregate forecasting accuracy over history-only and action-prefix baselines. Ablation and parameter analyses substantiate the effectiveness of latent modulation and contrastive supervision, while action consistency analysis further reveals stronger action-type discrimination and more coherent intensity-dependent responses.

\bibliographystyle{ACM-Reference-Format}
\bibliography{main}

@article{m5data,
  author = {Makridakis, Spyros and Spiliotis, Evangelos and Assimakopoulos, Vassilios},
  title = {The {M5} competition: Background, organization, and implementation},
  journal = {International Journal of Forecasting},
  year = {2022},
  volume = {38},
  number = {4},
  pages = {1325--1336},
  doi = {10.1016/j.ijforecast.2021.07.007},
  url = {https://doi.org/10.1016/j.ijforecast.2021.07.007}
}

@misc{mofdata,
  author = {{Ministry of Finance, Japan}},
  title = {Foreign Exchange Intervention Operations},
  howpublished = {Historical intervention data},
  url = {https://www.mof.go.jp/english/policy/international_policy/reference/feio/foreign_exchange_intervention_operations.csv},
  urldate = {2026-10-25}
}

@misc{fomcdata,
  author = {{Board of Governors of the Federal Reserve System}},
  title = {Meeting Calendars, Statements, and Minutes},
  howpublished = {Federal Open Market Committee},
  url = {https://www.federalreserve.gov/monetarypolicy/fomccalendars.htm},
  urldate = {2026-10-25}
}

@article{causalchambersdata,
  author = {Gamella, Juan L. and Peters, Jonas and B{\"u}hlmann, Peter},
  title = {Causal chambers as a real-world physical testbed for {AI} methodology},
  journal = {Nature Machine Intelligence},
  year = {2025},
  volume = {7},
  number = {1},
  pages = {107--118},
  doi = {10.1038/s42256-024-00964-x},
  url = {https://doi.org/10.1038/s42256-024-00964-x}
}

@article{daisydata,
  author = {Bittanti, S. and Piroddi, L.},
  title = {Nonlinear identification and control of a heat exchanger: A neural network approach},
  journal = {Journal of the Franklin Institute},
  year = {1997},
  volume = {334},
  number = {1},
  pages = {135--153},
  doi = {10.1016/S0016-0032(96)00059-2},
  url = {https://doi.org/10.1016/S0016-0032(96)00059-2}
}

@misc{shanghaidata,
  author = {Zhu, Jinhao},
  title = {Diabetes Datasets, {ShanghaiT1DM} and {ShanghaiT2DM}},
  year = {2022},
  publisher = {figshare},
  doi = {10.6084/m9.figshare.20425518.v5},
  url = {https://figshare.com/articles/dataset/Diabetes_Datasets/20425518/5}
}

@misc{hupadata,
  author = {Hidalgo, J. Ignacio and Alvarado, Jorge and Botella, Marta and Aramendi, Aranzazu and Velasco, J. Manuel and Garnica, Oscar},
  title = {{HUPA-UCM Diabetes Dataset}},
  year = {2024},
  publisher = {Mendeley Data},
  version = {1},
  doi = {10.17632/3hbcscwz44.1},
  url = {https://data.mendeley.com/datasets/3hbcscwz44/1}
}

@article{qwen3embedding,
  title = {{Qwen3 Embedding}: Advancing Text Embedding and Reranking Through Foundation Models},
  author = {Zhang, Yanzhao and Li, Mingxin and Long, Dingkun and Zhang, Xin and Lin, Huan and Yang, Baosong and Xie, Pengjun and Yang, An and Liu, Dayiheng and Lin, Junyang and Huang, Fei and Zhou, Jingren},
  journal = {arXiv preprint arXiv:2506.05176},
  year = {2025},
  url = {https://arxiv.org/abs/2506.05176}
}

@misc{gpt6luna2026,
  author = {{OpenAI}},
  title = {{GPT-6 Luna}},
  year = {2026},
  howpublished = {OpenAI API model documentation},
  url = {https://developers.openai.com/api/docs/models/gpt-6-luna},
  note = {Accessed October 14, 2026}
}

@misc{timesfm32026,
  title = {TimesFM-3: A zero-shot foundation model for multivariate forecasting},
  author = {Ayush Jain and Rajat Sen},
  year = {2026},
  month = aug,
  howpublished = {Google Research Blog},
  url = {https://research.google/blog/timesfm-3-a-zero-shot-foundation-model-for-multivariate-forecasting/},
  note = {Published August 31, 2026; accessed October 9, 2026}
}

@misc{chronos22025,
  title = {Chronos-2: From Univariate to Universal Forecasting},
  author = {Abdul Fatir Ansari and Oleksandr Shchur and Jaris Küken and Andreas Auer and Boran Han and Pedro Mercado and Syama Sundar Rangapuram and Huibin Shen and Lorenzo Stella and Xiyuan Zhang and Mononito Goswami and Shubham Kapoor and Danielle C. Maddix and Pablo Guerron and Tony Hu and Junming Yin and Nick Erickson and Prateek Mutalik Desai and Hao Wang and Huzefa Rangwala and George Karypis and Yuyang Wang and Michael Bohlke-Schneider},
  year = {2025},
  eprint = {2510.15821},
  archiveprefix = {arXiv},
  primaryclass = {cs.LG},
  url = {https://arxiv.org/abs/2510.15821}
}

@misc{toto22026,
  title = {Toto 2.0: Time Series Forecasting Enters the Scaling Era},
  author = {Emaad Khwaja and Chris Lettieri and Gerald Woo and Eden Belouadah and Marc Cenac and Guillaume Jarry and Enguerrand Paquin and Xunyi Zhao and Viktoriya Zhukov and Othmane Abou-Amal and Chenghao Liu and Ameet Talwalkar and David Asker},
  year = {2026},
  eprint = {2605.20119},
  archiveprefix = {arXiv},
  primaryclass = {cs.LG},
  url = {https://arxiv.org/abs/2605.20119}
}

@misc{moirai22025,
  title = {{Moirai 2.0}: When Less Is More for Time Series Forecasting},
  author = {Chenghao Liu and Taha Aksu and Juncheng Liu and Xu Liu and Hanshu Yan and Quang Pham and Silvio Savarese and Doyen Sahoo and Caiming Xiong and Junnan Li},
  year = {2025},
  eprint = {2511.11698},
  archiveprefix = {arXiv},
  primaryclass = {cs.LG},
  url = {https://arxiv.org/abs/2511.11698}
}

@inproceedings{timer2024,
  title = {Timer: Generative Pre-trained Transformers Are Large Time Series Models},
  author = {Liu, Yong and Zhang, Haoran and Li, Chenyu and Huang, Xiangdong and Wang, Jianmin and Long, Mingsheng},
  booktitle = {Proceedings of the 41st International Conference on Machine Learning},
  pages = {32369--32399},
  year = {2024},
  editor = {Salakhutdinov, Ruslan and Kolter, Zico and Heller, Katherine and Weller, Adrian and Oliver, Nuria and Scarlett, Jonathan and Berkenkamp, Felix},
  volume = {235},
  series = {Proceedings of Machine Learning Research},
  month = {21--27 Jul},
  publisher = {PMLR},
  url = {https://proceedings.mlr.press/v235/liu24cb.html}
}

@inproceedings{timemoe2025,
  author = {Shi, Xiaoming and Wang, Shiyu and Nie, Yuqi and Li, Dianqi and Ye, Zhou and Wen, Qingsong and Jin, Ming},
  booktitle = {International Conference on Learning Representations},
  editor = {Y. Yue and A. Garg and N. Peng and F. Sha and R. Yu},
  pages = {34635--34667},
  title = {Time-MoE: Billion-Scale Time Series Foundation Models with Mixture of Experts},
  url = {https://proceedings.iclr.cc/paper_files/paper/2025/file/558d48c1f08675daa636e09bfe94a89e-Paper-Conference.pdf},
  volume = {2025},
  year = {2025}
}

@inproceedings{niextreme2026,
  title = {{UniExtreme}: A Universal Foundation Model for Extreme Weather Forecasting},
  author = {Ni, Hang and Zhang, Weijia and Liu, Hao},
  booktitle = {Proceedings of the 32nd ACM SIGKDD Conference on Knowledge Discovery and Data Mining V.1},
  publisher = {ACM},
  year = {2026},
  month = apr,
  pages = {1101--1112},
  doi = {10.1145/3770854.3780172},
  url = {https://doi.org/10.1145/3770854.3780172}
}

@article{promptcast2024,
  title = {PromptCast: A New Prompt-Based Learning Paradigm for Time Series Forecasting},
  volume = {36},
  issn = {2326-3865},
  url = {http://dx.doi.org/10.1109/TKDE.2023.3342137},
  doi = {10.1109/tkde.2023.3342137},
  number = {11},
  journal = {IEEE Transactions on Knowledge and Data Engineering},
  publisher = {Institute of Electrical and Electronics Engineers (IEEE)},
  author = {Xue, Hao and Salim, Flora D.},
  year = {2024},
  month = nov,
  pages = {6851--6864}
}

@article{timecma2025,
  title = {TimeCMA: Towards LLM-Empowered Multivariate Time Series Forecasting via Cross-Modality Alignment},
  volume = {39},
  issn = {2159-5399},
  url = {http://dx.doi.org/10.1609/aaai.v39i18.34067},
  doi = {10.1609/aaai.v39i18.34067},
  number = {18},
  journal = {Proceedings of the AAAI Conference on Artificial Intelligence},
  publisher = {Association for the Advancement of Artificial Intelligence (AAAI)},
  author = {Liu, Chenxi and Xu, Qianxiong and Miao, Hao and Yang, Sun and Zhang, Lingzheng and Long, Cheng and Li, Ziyue and Zhao, Rui},
  year = {2025},
  month = apr,
  pages = {18780--18788}
}

@inproceedings{dualsg2025,
  series = {MM ’25},
  title = {DualSG: A Dual-Stream Explicit Semantic-Guided Multivariate Time Series Forecasting Framework},
  url = {http://dx.doi.org/10.1145/3746027.3755458},
  doi = {10.1145/3746027.3755458},
  booktitle = {Proceedings of the 33rd ACM International Conference on Multimedia},
  publisher = {ACM},
  author = {Ding, Kuiye and Fan, Fanda and Wang, Yao and Jian, Ruijie and Wang, Xiaorui and Gong, Luqi and Jiang, Yishan and Luo, Chunjie and Zhan, Jianfeng},
  year = {2025},
  month = oct,
  pages = {508--517}
}

@inproceedings{timellm2024,
  author = {Jin, Ming and Wang, Shiyu and Ma, Lintao and Chu, Zhixuan and Zhang, James and Shi, Xiaoming and Chen, Pin-Yu and Liang, Yuxuan and Li, Yuan-Fang and Pan, Shirui and Wen, Qingsong},
  booktitle = {International Conference on Learning Representations},
  editor = {B. Kim and Y. Yue and S. Chaudhuri and K. Fragkiadaki and M. Khan and Y. Sun},
  pages = {23857--23880},
  title = {Time-LLM: Time Series Forecasting by Reprogramming Large Language Models},
  url = {https://proceedings.iclr.cc/paper_files/paper/2024/file/680b2a8135b9c71278a09cafb605869e-Paper-Conference.pdf},
  volume = {2024},
  year = {2024}
}

@inproceedings{unitime2024,
  series = {WWW ’24},
  title = {UniTime: A Language-Empowered Unified Model for Cross-Domain Time Series Forecasting},
  url = {http://dx.doi.org/10.1145/3589334.3645434},
  doi = {10.1145/3589334.3645434},
  booktitle = {Proceedings of the ACM Web Conference 2024},
  publisher = {ACM},
  author = {Liu, Xu and Hu, Junfeng and Li, Yuan and Diao, Shizhe and Liang, Yuxuan and Hooi, Bryan and Zimmermann, Roger},
  year = {2024},
  month = may,
  pages = {4095--4106}
}

@inproceedings{timevlm2025,
  title = {Time-{VLM}: Exploring Multimodal Vision-Language Models for Augmented Time Series Forecasting},
  author = {Zhong, Siru and Ruan, Weilin and Jin, Ming and Li, Huan and Wen, Qingsong and Liang, Yuxuan},
  booktitle = {Proceedings of the 42nd International Conference on Machine Learning},
  pages = {78478--78497},
  year = {2025},
  editor = {Singh, Aarti and Fazel, Maryam and Hsu, Daniel and Lacoste-Julien, Simon and Berkenkamp, Felix and Maharaj, Tegan and Wagstaff, Kiri and Zhu, Jerry},
  volume = {267},
  series = {Proceedings of Machine Learning Research},
  month = {13--19 Jul},
  publisher = {PMLR},
  url = {https://proceedings.mlr.press/v267/zhong25a.html}
}

@inproceedings{timemmd2024,
  author = {Liu, Haoxin and Xu, Shangqing and Zhao, Zhiyuan and Kong, Lingkai and Kamarthi, Harshavardhan and Sasanur, Aditya B. and Sharma, Megha and Cui, Jiaming and Wen, Qingsong and Zhang, Chao and Prakash, B. Aditya},
  booktitle = {Advances in Neural Information Processing Systems},
  doi = {10.52202/079017-2476},
  editor = {A. Globerson and L. Mackey and D. Belgrave and A. Fan and U. Paquet and J. Tomczak and C. Zhang},
  pages = {77888--77933},
  publisher = {Curran Associates, Inc.},
  title = {Time-MMD: Multi-Domain Multimodal Dataset for Time Series Analysis},
  url = {https://proceedings.neurips.cc/paper_files/paper/2024/file/8e7768122f3eeec6d77cd2b424b72413-Paper-Datasets_and_Benchmarks_Track.pdf},
  volume = {37},
  year = {2024}
}

@inproceedings{wang2024from,
  author = {Wang, Xinlei and Feng, Maike and Qiu, Jing and Gu, Jinjin and Zhao, Junhua},
  booktitle = {Advances in Neural Information Processing Systems},
  doi = {10.52202/079017-1853},
  editor = {A. Globerson and L. Mackey and D. Belgrave and A. Fan and U. Paquet and J. Tomczak and C. Zhang},
  pages = {58118--58153},
  publisher = {Curran Associates, Inc.},
  title = {From News to Forecast: Integrating Event Analysis in LLM-Based Time Series Forecasting with Reflection},
  url = {https://proceedings.neurips.cc/paper_files/paper/2024/file/6aef8bffb372096ee73d98da30119f89-Paper-Conference.pdf},
  volume = {37},
  year = {2024}
}

@article{chattime2025,
  title = {ChatTime: A Unified Multimodal Time Series Foundation Model Bridging Numerical and Textual Data},
  volume = {39},
  issn = {2159-5399},
  url = {http://dx.doi.org/10.1609/aaai.v39i12.33384},
  doi = {10.1609/aaai.v39i12.33384},
  number = {12},
  journal = {Proceedings of the AAAI Conference on Artificial Intelligence},
  publisher = {Association for the Advancement of Artificial Intelligence (AAAI)},
  author = {Wang, Chengsen and Qi, Qi and Wang, Jingyu and Sun, Haifeng and Zhuang, Zirui and Wu, Jinming and Zhang, Lei and Liao, Jianxin},
  year = {2025},
  month = apr,
  pages = {12694--12702}
}

@inproceedings{cik2025,
  title = {Context is Key: A Benchmark for Forecasting with Essential Textual Information},
  author = {Williams, Andrew Robert and Ashok, Arjun and Marcotte, \'{E}tienne and Zantedeschi, Valentina and Subramanian, Jithendaraa and Riachi, Roland and Requeima, James and Lacoste, Alexandre and Rish, Irina and Chapados, Nicolas and Drouin, Alexandre},
  booktitle = {Proceedings of the 42nd International Conference on Machine Learning},
  pages = {66887--66944},
  year = {2025},
  editor = {Singh, Aarti and Fazel, Maryam and Hsu, Daniel and Lacoste-Julien, Simon and Berkenkamp, Felix and Maharaj, Tegan and Wagstaff, Kiri and Zhu, Jerry},
  volume = {267},
  series = {Proceedings of Machine Learning Research},
  month = {13--19 Jul},
  publisher = {PMLR},
  url = {https://proceedings.mlr.press/v267/williams25a.html}
}

@inproceedings{camef2025,
  series = {KDD ’25},
  title = {CAMEF: Causal-Augmented Multi-Modality Event-Driven Financial Forecasting by Integrating Time Series Patterns and Salient Macroeconomic Announcements},
  url = {http://dx.doi.org/10.1145/3711896.3736872},
  doi = {10.1145/3711896.3736872},
  booktitle = {Proceedings of the 31st ACM SIGKDD Conference on Knowledge Discovery and Data Mining V.2},
  publisher = {ACM},
  author = {Zhang, Yang and Yang, Wenbo and Wang, Jun and Ma, Qiang and Xiong, Jie},
  year = {2025},
  month = aug,
  pages = {3867--3878}
}

@inproceedings{aurora2026,
  author = {Wu, Xingjian and Jin, Jianxin and Qiu, Wanghui and Chen, Peng and Shu, Yang and Yang, Bin and Guo, Chenjuan},
  booktitle = {International Conference on Learning Representations},
  editor = {C. Vondrick and B. Hariharan and C. Raffel and L. Pinto and D. Yang and A. Faust},
  pages = {103239--103277},
  title = {Aurora: Towards Universal Generative Multimodal Time Series Forecasting},
  url = {https://proceedings.iclr.cc/paper_files/paper/2026/file/a8e18adf6489fd3a27417d128660eaa9-Paper-Conference.pdf},
  volume = {2026},
  year = {2026}
}

@misc{qiyaom2026,
  title = {{QiYao-M}: Multimodal Time Series Foundation Model with Role-Aware Modeling of Endogenous and Exogenous Modalities},
  author = {Cheng, Hanyin and Wang, Linfeng and Qu, Zhengbo and Shu, Yang and Rao, Zhongwen and Wang, Meng and Li, Yijie and Jiang, Xin and Yang, Bin and Guo, Chenjuan},
  year = {2026},
  eprint = {2609.34842},
  archivePrefix = {arXiv},
  primaryClass = {cs.LG},
  url = {https://arxiv.org/abs/2609.34842}
}

@inproceedings{tadiff2026,
  title = {What if Tomorrow is the World Cup Final? {C}ounterfactual Time Series Forecasting with Textual Conditions},
  author = {Gu, Shuqi and Zhao, Yongxiang and Jing, Baoyu and Ren, Kan},
  booktitle = {Proceedings of the 43rd International Conference on Machine Learning},
  pages = {37376--37394},
  year = {2026},
  editor = {Zhang, Tong and Dudik, Miroslav and Jaggi, Martin and Agarwal, Alekh and Li, Sharon and Schuurmans, Dale and Zhu, Jerry and Berkenkamp, Felix and Dong, Hanze and Bietti, Alberto},
  volume = {306},
  series = {Proceedings of Machine Learning Research},
  month = {06--11 Jul},
  publisher = {PMLR},
  url = {https://proceedings.mlr.press/v306/gu26r.html}
}

@misc{pythia2026,
  title = {Pythia: Toward Foundation World Models for Multimodal Time Series},
  author = {Xilin Dai and Hongzhou Chen and Yifan Hu and Yiding Liu and Zewei Dong and Jiang-Ming Yang},
  year = {2026},
  eprint = {2610.05240},
  archiveprefix = {arXiv},
  primaryclass = {cs.LG},
  url = {https://arxiv.org/abs/2610.05240}
}

@misc{worldts2026,
  title = {WorldTS: World Modeling for Multimodal Covariate-aware Time Series Forecasting},
  author = {Yuhan Zhu and Xiangfei Qiu and Hanyin Cheng and Wangmeng Shen and Chenjuan Guo and Bin Yang and Jilin Hu and Christian S. Jensen},
  year = {2026},
  eprint = {2609.31162},
  archiveprefix = {arXiv},
  primaryclass = {cs.LG},
  url = {https://arxiv.org/abs/2609.31162}
}

@inproceedings{timecapsule2025,
  series = {KDD ’25},
  title = {TimeCapsule: Solving the Jigsaw Puzzle of Long-Term Time Series Forecasting with Compressed Predictive Representations},
  url = {http://dx.doi.org/10.1145/3711896.3737157},
  doi = {10.1145/3711896.3737157},
  booktitle = {Proceedings of the 31st ACM SIGKDD Conference on Knowledge Discovery and Data Mining V.2},
  publisher = {ACM},
  author = {Lu, Yihang and Xu, Yangyang and Qin, Qitao and Meng, Xianwei},
  year = {2025},
  month = aug,
  pages = {1987--1998}
}

@inproceedings{latenttsf2026,
  title = {From Observations to States: Latent Time Series Forecasting},
  author = {Yang, Jie and Hu, Yifan and Li, Yuante and Zhang, Kexin and Ding, Kaize and Yu, Philip S.},
  booktitle = {Proceedings of the 43rd International Conference on Machine Learning},
  pages = {144796--144819},
  year = {2026},
  editor = {Zhang, Tong and Dudik, Miroslav and Jaggi, Martin and Agarwal, Alekh and Li, Sharon and Schuurmans, Dale and Zhu, Jerry and Berkenkamp, Felix and Dong, Hanze and Bietti, Alberto},
  volume = {306},
  series = {Proceedings of Machine Learning Research},
  month = {06--11 Jul},
  publisher = {PMLR},
  url = {https://proceedings.mlr.press/v306/yang26y.html}
}

@inproceedings{ct2022,
  title = {Causal Transformer for Estimating Counterfactual Outcomes},
  author = {Melnychuk, Valentyn and Frauen, Dennis and Feuerriegel, Stefan},
  booktitle = {Proceedings of the 39th International Conference on Machine Learning},
  pages = {15293--15329},
  year = {2022},
  editor = {Chaudhuri, Kamalika and Jegelka, Stefanie and Song, Le and Szepesvari, Csaba and Niu, Gang and Sabato, Sivan},
  volume = {162},
  series = {Proceedings of Machine Learning Research},
  month = {17--23 Jul},
  publisher = {PMLR},
  url = {https://proceedings.mlr.press/v162/melnychuk22a.html}
}

@inproceedings{mambacdsp2025,
  author = {Wang, Haotian and Li, Haoxuan and Zou, Hao and Chi, Haoang and Lan, Long and Huang, Wanrong and Yang, Wenjing},
  booktitle = {International Conference on Learning Representations},
  editor = {Y. Yue and A. Garg and N. Peng and F. Sha and R. Yu},
  pages = {15000--15025},
  title = {Effective and Efficient Time-Varying Counterfactual Prediction with State-Space Models},
  url = {https://proceedings.iclr.cc/paper_files/paper/2025/file/26f16a11f4dbba3eef876b571d9f3200-Paper-Conference.pdf},
  volume = {2025},
  year = {2025}
}

@inproceedings{meddreamer2026,
  title = {medDreamer: Model-Based Reinforcement Learning with Latent Imagination on Complex EHRs for Clinical Decision Support},
  url = {http://dx.doi.org/10.1145/3770854.3780292},
  doi = {10.1145/3770854.3780292},
  booktitle = {Proceedings of the 32nd ACM SIGKDD Conference on Knowledge Discovery and Data Mining V.1},
  publisher = {ACM},
  author = {Xu, Qianyi and Habib, Gousia and Wu, Feng and Perera, Dilruk and Feng, Mengling},
  year = {2026},
  month = apr,
  pages = {1693--1704}
}

@misc{mobiwm2026,
  title = {Beyond Static Forecasting: Unleashing the Power of World Models for Mobile Traffic Extrapolation},
  author = {Xiaoqian Qi and Haoye Chai and Yue Wang and Yong Li},
  year = {2026},
  eprint = {2604.08199},
  archiveprefix = {arXiv},
  primaryclass = {cs.NI},
  url = {https://arxiv.org/abs/2604.08199}
}

@inproceedings{cedar2026,
  title = {CEDAR: Controlled and Event-Driven Demand Forecasting via Residual Decomposition},
  url = {http://dx.doi.org/10.1145/3770855.3818338},
  doi = {10.1145/3770855.3818338},
  booktitle = {Proceedings of the 32nd ACM SIGKDD Conference on Knowledge Discovery and Data Mining V.2},
  publisher = {ACM},
  author = {Meng, Junjie and Zhang, Ranxu and Zhang, Zi-an and Liu, Shujun and Qi, Xiaoning and Xu, Xiaozhou and Zhang, Yanyong and Xiong, Hui and Wang, Chao},
  year = {2026},
  month = aug,
  pages = {7797--7808}
}

@misc{zhang2026on,
  title = {On the Divergence of Accuracy and Mechanism Consistency in Time Series World Models},
  author = {Haochen Zhang and Jiaheng Guo and Zhen Xu and Zachary Plotkin and Nicholas Konz and Zhen Tan and Tianlong Chen},
  year = {2026},
  eprint = {2610.01842},
  archiveprefix = {arXiv},
  primaryclass = {cs.AI},
  url = {https://arxiv.org/abs/2610.01842}
}

@misc{reasoncast2026,
  title = {ReasonCast: Agentic Demand Forecasting with Selective Semantic Reasoning},
  author = {Ziyue Yang and Chaolin Xu and Yijing Wang and Tiankai Gu and Hui Yang and Yanhong Lin and Kaiyuan Liu and Fei Xiao},
  year = {2026},
  eprint = {2608.15291},
  archiveprefix = {arXiv},
  primaryclass = {cs.AI},
  url = {https://arxiv.org/abs/2608.15291}
}

@inproceedings{das2024decoder,
  title = {A decoder-only foundation model for time-series forecasting},
  author = {Das, Abhimanyu and Kong, Weihao and Sen, Rajat and Zhou, Yichen},
  booktitle = {Proceedings of the 41st International Conference on Machine Learning},
  pages = {10148--10167},
  volume = {235},
  series = {Proceedings of Machine Learning Research},
  publisher = {PMLR},
  year = {2024},
  url = {https://proceedings.mlr.press/v235/das24c.html}
}

@article{hafner2025mastering,
  title = {Mastering diverse control tasks through world models},
  author = {Hafner, Danijar and Pasukonis, Jurgis and Ba, Jimmy and Lillicrap, Timothy},
  journal = {Nature},
  volume = {640},
  pages = {647--653},
  year = {2025},
  doi = {10.1038/s41586-025-08744-2},
  url = {https://www.nature.com/articles/s41586-025-08744-2}
}

@inproceedings{genie2024,
  title = {Genie: Generative Interactive Environments},
  author = {Bruce, Jake and Dennis, Michael D and Edwards, Ashley and Parker-Holder, Jack and Shi, Yuge and Hughes, Edward and Lai, Matthew and Mavalankar, Aditi and Steigerwald, Richie and Apps, Chris and Aytar, Yusuf and Bechtle, Sarah Maria Elisabeth and Behbahani, Feryal and Chan, Stephanie C.Y. and Heess, Nicolas and Gonzalez, Lucy and Osindero, Simon and Ozair, Sherjil and Reed, Scott and Zhang, Jingwei and Zolna, Konrad and Clune, Jeff and Freitas, Nando De and Singh, Satinder and Rockt\"{a}schel, Tim},
  booktitle = {Proceedings of the 41st International Conference on Machine Learning},
  pages = {4603--4623},
  volume = {235},
  series = {Proceedings of Machine Learning Research},
  publisher = {PMLR},
  year = {2024},
  url = {https://proceedings.mlr.press/v235/bruce24a.html}
}

@misc{vjepa2024,
  title = {Revisiting Feature Prediction for Learning Visual Representations from Video},
  author = {Adrien Bardes and Quentin Garrido and Jean Ponce and Xinlei Chen and Michael Rabbat and Yann LeCun and Mahmoud Assran and Nicolas Ballas},
  year = {2024},
  eprint = {2404.08471},
  archiveprefix = {arXiv},
  primaryclass = {cs.CV},
  url = {https://arxiv.org/abs/2404.08471}
}

@article{qwenrobot-world,
  title = {Qwen-RobotWorld Technical Report: Unifying Embodied World Modeling through Language-Conditioned Video Generation},
  author = {Qwen Team},
  year = {2026}
}

@InProceedings{pmlr-v305-jang25a,
  title = {DreamGen: Unlocking Generalization in Robot Learning through Video World Models},
  author = {Jang, Joel and Ye, Seonghyeon and Lin, Zongyu and Xiang, Jiannan and Bjorck, Johan and Fang, Yu and Hu, Fengyuan and Huang, Spencer and Kundalia, Kaushil and Lin, Yen-Chen and Magne, Lo\"{i}c and Mandlekar, Ajay and Narayan, Avnish and Tan, You Liang and Wang, Guanzhi and Wang, Jing and Wang, Qi and Xu, Yinzhen and Zeng, Xiaohui and Zheng, Kaiyuan and Zheng, Ruijie and Liu, Ming-Yu and Zettlemoyer, Luke and Fox, Dieter and Kautz, Jan and Reed, Scott and Zhu, Yuke and Fan, Linxi},
  booktitle = {Proceedings of The 9th Conference on Robot Learning},
  pages = {5170--5194},
  year = {2025},
  editor = {Lim, Joseph and Song, Shuran and Park, Hae-Won},
  volume = {305},
  series = {Proceedings of Machine Learning Research},
  month = {27--30 Sep},
  publisher = {PMLR},
  url = {https://proceedings.mlr.press/v305/jang25a.html}
}

@article{salinas2020deepar,
  title = {{DeepAR}: Probabilistic forecasting with autoregressive recurrent networks},
  author = {Salinas, David and Flunkert, Valentin and Gasthaus, Jan and Januschowski, Tim},
  journal = {International Journal of Forecasting},
  volume = {36},
  number = {3},
  pages = {1181--1191},
  year = {2020},
  doi = {10.1016/j.ijforecast.2019.07.001},
  url = {https://doi.org/10.1016/j.ijforecast.2019.07.001}
}

@inproceedings{lazaris2019lstm,
  title = {An {LSTM} Framework For Modeling Network Traffic},
  author = {Lazaris, Aggelos and Prasanna, Viktor K.},
  booktitle = {2019 IFIP/IEEE Symposium on Integrated Network and Service Management (IM)},
  publisher = {IEEE},
  year = {2019},
  url = {https://ieeexplore.ieee.org/document/8717920/}
}

@article{zhang2019deeplob,
  title = {{DeepLOB}: Deep Convolutional Neural Networks for Limit Order Books},
  author = {Zhang, Zihao and Zohren, Stefan and Roberts, Stephen},
  journal = {IEEE Transactions on Signal Processing},
  volume = {67},
  number = {11},
  pages = {3001--3012},
  year = {2019},
  doi = {10.1109/TSP.2019.2907260},
  url = {https://doi.org/10.1109/TSP.2019.2907260}
}

@misc{nimadi2026,
  title = {From Consistency to Complementarity: Aligned and Disentangled Multi-modal Learning for Time Series Understanding and Reasoning},
  author = {Hang Ni and Weijia Zhang and Fei Wang and Zezhi Shao and Hao Liu},
  year = {2026},
  eprint = {2601.21436},
  archiveprefix = {arXiv},
  primaryclass = {cs.LG},
  url = {https://arxiv.org/abs/2601.21436}
}

@inproceedings{zha2023rank,
  title = {{Rank-N-Contrast}: Learning Continuous Representations for Regression},
  author = {Zha, Kaiwen and Cao, Peng and Son, Jeany and Yang, Yuzhe and Katabi, Dina},
  booktitle = {Advances in Neural Information Processing Systems},
  volume = {36},
  year = {2023},
  url = {https://papers.nips.cc/paper_files/paper/2023/hash/39e9c5913c970e3e49c2df629daff636-Abstract-Conference.html}
}

\appendix

\section{Baseline Implementation Details}
\label{app:baseline-details}

MM-TSFlib~\cite{timemmd2024} uses iTransformer for temporal modeling and GPT-2 for text encoding.
CovTSWM~\cite{zhang2026on} combines TiDE with a frozen autoencoder and output gating to condition forecasts on structured actions.
We adapt its covariate schema to each dataset's recorded controls and align their values, magnitudes, and event indicators with the forecast timeline.
For FOMC, these covariates encode the explicitly announced target interest rate when available.

\section{Evaluation Metric Details}
\label{app:metrics}

For evaluation across test instances, we add an instance index $\nu$ to the notation in \secref{sec:task}, yielding history $\mathbf{x}_\nu$, target $\mathbf{y}_\nu$, and prediction $\widehat{\mathbf{y}}_\nu$.
For each instance, $L$ and $H$ denote its history length and forecast horizon.
Here, $t$ locally indexes chronological history positions in $x_{\nu,t}$ ($t=1,\ldots,L$) and forecast steps in $y_{\nu,t}$ and $\widehat{y}_{\nu,t}$ ($t=1,\ldots,H$).
The comma separates instance and temporal indices.
The prediction errors are
\begin{equation*}
\mathrm{MAE}_\nu = \frac{1}{H}
    \sum_{t=1}^{H}|\widehat{y}_{\nu,t}-y_{\nu,t}|,\
\mathrm{RMSE}_\nu = \sqrt{\frac{1}{H}
    \sum_{t=1}^{H}(\widehat{y}_{\nu,t}-y_{\nu,t})^2}.
\end{equation*}
The history mean $\mu_{\mathbf{x}_\nu}$, population standard deviation $\sigma_{\mathbf{x}_\nu}$, and mean absolute lag-one difference $\mathrm{MAE}^{\mathrm{naive}}_\nu$ are computed separately from each instance's history:
\begin{equation*}
\begin{aligned}
&\mu_{\mathbf{x}_\nu} = \frac{1}{L}
    \sum_{t=1}^{L}x_{\nu,t},\ \
\sigma_{\mathbf{x}_\nu} = \sqrt{\frac{1}{L}
    \sum_{t=1}^{L}(x_{\nu,t}-\mu_{\mathbf{x}_\nu})^2},\\
&\mathrm{MAE}^{\mathrm{naive}}_\nu = \frac{1}{L-1}
    \sum_{t=2}^{L}|x_{\nu,t}-x_{\nu,t-1}|.
\end{aligned}
\end{equation*}
These expressions assume fully observed windows. With missing values, each sum includes only observed targets, observed history values, or adjacent history pairs with both values originally observed, as appropriate, and its denominator is replaced by the corresponding valid count.
The normalized and scaled errors are
$\mathrm{nMAE}_\nu = \mathrm{MAE}_\nu/\sigma_{\mathbf{x}_\nu},\
\mathrm{nRMSE}_\nu = \mathrm{RMSE}_\nu/\sigma_{\mathbf{x}_\nu},\
\mathrm{MASE}_\nu = \mathrm{MAE}_\nu/\mathrm{MAE}^{\mathrm{naive}}_\nu$, where MASE uses the nonseasonal naive scale computed from each instance's history.
Instances with no valid observations or a zero normalization scale are excluded from the corresponding metric average.
We average each metric over valid test instances for each $(L,H)$ configuration, taking the RMSE square root before averaging.
Dataset-level results average configurations within each dataset. Overall results weight all configurations across datasets equally.

\section{Counterfactual Action Construction Details}
\label{app:action-construction}

Each generation request to the LLM includes the factual action and specifies which attribute may change, so type alternatives vary the operation while intensity alternatives retain the operation and follow the assigned magnitude ordering.
All candidates preserve unrelated context, units, and temporal information to keep the contrast focused on the intended action variation.
The consistent and regular action descriptions in \bench\xspace facilitate checks of format, required fields, and the requested variations.
Programmatic validation of all generated actions, supplemented by manual inspection of sampled examples, revealed no formatting errors.
However, handling irregular, unconstrained action text remains a nontrivial challenge beyond the scope of this paper, motivating future benchmarks and solutions for such inputs.

\section{Computational Complexity}
\label{app:complexity}

\model\xspace augments a frozen TSFM with one context token and lightweight latent modulation.
Beyond text encoding and native TSFM computation, adapters and PLM cost $O\bigl(N_{\mathrm{a}}D^2\allowbreak{}+N_{\mathrm{f}}(D^2\allowbreak{}+\allowbreak{}N_{\mathrm{a}}D\allowbreak{}+DR\allowbreak{}+N_{\mathrm{p}}R)\bigr)$ per forecast, assuming projection widths of $O(D)$.
Reduced QR uses PyTorch's built-in differentiable \texttt{torch.linalg.qr} and costs $O(DR^2)$, while prototype projection costs $O(N_{\mathrm{p}}DR)$ given reusable frozen reference encodings.
Both can be cached at inference.
With $R=32$ and $N_{\mathrm{p}}=24$, PLM remains small and linear in $N_{\mathrm{f}}$ for fixed action length.
Training additionally evaluates $1+N_{\mathrm{t}}+N_{\mathrm{s}}$ action-conditioned branches for ACL.
After pooling and projection, type matching costs $O((N_{\mathrm{t}}+1)^2D)$, while intensity ranking costs $O((N_{\mathrm{s}}+1)^2D+(N_{\mathrm{s}}+1)^3)$ under direct evaluation of the comparison-set denominators.
The cubic term is limited to six intensity levels because $N_{\mathrm{t}}=N_{\mathrm{s}}=5$.
These extra branches and contrastive losses are training-only and incur no inference cost.

\section{Dataset Details}
\label{app:datasets}

The history lengths and forecast horizons reflect each domain's sampling frequency and action timescale, as summarized in \tabref{tab:dataset-splits}.

\noindent\textbf{M5.}
M5~\cite{m5data} contains daily item--store sales and weekly average prices.
Actions describe recorded price changes, including the previous and updated prices and their effective interval.
We use $L=112$ days and $H\in\{7,28\}$ days to capture weekly and four-week demand responses, with a fixed 1\% sample shared by all methods.

\noindent\textbf{MOF.}
MOF~\cite{mofdata} pairs Japan's foreign-exchange interventions with yen-per-dollar observations from FRED.
Actions specify the traded currencies, direction, and amount, combining operations on the same date.
We use $L=60$ market observations and $H\in\{1,5\}$ to examine immediate and short-term responses, aligning observations to separate pre-intervention history from subsequent outcomes.

\noindent\textbf{FOMC.}
FOMC~\cite{fomcdata} pairs monetary policy statements with U.S. two-year Treasury yields.
The full statement serves as the textual action, retaining policy information beyond the announced rate.
We use $L=60$ and $H\in\{1,5\}$ market observations, with targets beginning after the announcement date.

\noindent\textbf{Causal Chambers.}
Causal Chambers~\cite{causalchambersdata} provides controlled wind-tunnel experiments that vary inlet fan load while holding other actuators fixed.
We forecast downstream absolute pressure with $(L,H)=(100,100)$.
Lengths count native observations because sampling intervals are irregular.

\noindent\textbf{DaISy.}
The DaISy heat-exchanger data~\cite{daisydata} record liquid flow and outlet temperature every second.
We use $L\in\{120,300\}$ and $H=1$, reflecting two or five minutes of history and the per-step changes in flow.
The target is the temperature at the current input step, with history ending at the preceding step.

\noindent\textbf{Shanghai Diabetes.}
Shanghai Diabetes~\cite{shanghaidata} combines 15-minute glucose measurements with records of meals, medication, and insulin administration.
We consolidate simultaneous actions while preserving food quantities and medication details.
We use $L=96$ and $H\in\{4,8\}$, corresponding to a 24-hour history and forecast horizons of one or two hours.
History excludes the glucose measurement contemporaneous with the action.

\noindent\textbf{HUPA-UCM.}
HUPA-UCM~\cite{hupadata} records glucose, carbohydrate intake, and insulin doses for people with type 1 diabetes.
We use the raw FreeStyle historical glucose records at approximately 15-minute intervals, rather than the released 5-minute interpolated series.
We set $L=96$ (approximately 24 hours) and $H\in\{4,8\}$.
We exclude instances with long sampling gaps and participants whose pump records cannot be parsed consistently.

\begin{table}[h]
\centering
\vspace{-5pt}
\caption{History--horizon configurations and split sizes.}
\vspace{-5pt}
\label{tab:dataset-splits}
\small
\begin{tabular}{l|c|rrr}
\toprule
\textbf{Dataset} & \textbf{\boldmath$(L,H)$} & \textbf{Train} & \textbf{Val.} & \textbf{Test} \\
\midrule
M5 & $(112,7)$ & 438 & 147 & 147 \\
M5 & $(112,28)$ & 437 & 142 & 147 \\
MOF & $(60,1)$ & 57 & 15 & 21 \\
MOF & $(60,5)$ & 57 & 15 & 21 \\
FOMC & $(60,1)$ & 117 & 39 & 40 \\
FOMC & $(60,5)$ & 116 & 39 & 40 \\
Causal Chambers & $(100,100)$ & 54 & 18 & 18 \\
DaISy & $(120,1)$ & 2208 & 656 & 776 \\
DaISy & $(300,1)$ & 1920 & 440 & 740 \\
Shanghai Diabetes & $(96,4)$ & 1687 & 487 & 678 \\
Shanghai Diabetes & $(96,8)$ & 1687 & 487 & 678 \\
HUPA-UCM & $(96,4)$ & 274 & 92 & 90 \\
HUPA-UCM & $(96,8)$ & 273 & 92 & 85 \\
\bottomrule
\end{tabular}
\end{table}

In terms of data ethics, all datasets were obtained from publicly available sources through their official distribution channels.
Our study uses existing records and involves no new data collection from human participants.

\begin{table}[h]
\centering
\caption{Comparison with covariate action conditioning.}
\vspace{-5pt}
\label{tab:covariate-conditioning}
\small
\setlength{\tabcolsep}{3pt}
\begin{tabular}{l|ccc}
\toprule
\textbf{Method} & \textbf{nMAE $\downarrow$} & \textbf{nRMSE $\downarrow$} & \textbf{MASE $\downarrow$} \\
\midrule
\textbf{Vanilla} & 1.0540 & 1.2087 & 3.5023 \\
\textbf{Covariates (zero-shot)} & 1.1097 & 1.2519 & 3.9601 \\
\textbf{Covariates (full tuning)} & 1.0207 & 1.1340 & 3.4507 \\
\textbf{Covariates (head + LoRA)} & 1.1035 & 1.2438 & 3.9267 \\
\midrule
\textbf{TSFM++} & \textbf{0.9983} & \textbf{1.1116} & \textbf{3.3634} \\
\bottomrule
\end{tabular}
\vspace{-5pt}
\end{table}

\section{Covariate Conditioning Study}
\label{app:covariate-conditioning}

We compare textual action modeling with TimesFM~3.0's native support for structured covariates.
Using the CovTSWM action encoding described in \appref{app:baseline-details}, we evaluate native covariate conditioning zero-shot and with either full-parameter or parameter-efficient fine-tuning on the same data splits.
Both fine-tuning variants start from pretrained TimesFM~3.0 weights: the former updates all parameters at a learning rate of $10^{-5}$, while the latter tunes the output head and rank-8 attention-layer LoRA at $10^{-6}$, keeping the remaining weights frozen.
Both use AdamW for up to ten epochs and early stopping after three epochs without validation improvement.
\tabref{tab:covariate-conditioning} shows that zero-shot covariate inputs increase aggregate errors relative to vanilla forecasting, while full fine-tuning improves all three metrics over vanilla.
Head and LoRA tuning yields only modest gains over zero-shot covariate conditioning and remains less accurate than vanilla.
Neither fine-tuning strategy closes the gap to TSFM++'s lightweight textual action adaptation.

\begin{table}[h]
\centering
\caption{Performance against variants that fine-tune TSFMs.}
\label{tab:backbone-adaptation}
\small
\setlength{\tabcolsep}{3pt}
\begin{tabular}{l|ccc}
\toprule
\textbf{Method} & \textbf{nMAE $\downarrow$} & \textbf{nRMSE $\downarrow$} & \textbf{MASE $\downarrow$} \\
\midrule
\textbf{\model\ (frozen TSFM)} & \textbf{0.8997} & \textbf{1.0185} & \textbf{3.0651} \\
\textbf{+ Full tuning} & 0.9082 & 1.0224 & 3.0916 \\
\textbf{+ Head and LoRA} & 0.9003 & 1.0188 & 3.0682 \\
\bottomrule
\end{tabular}
\end{table}

\section{TSFM Fine-Tuning Study}
\label{app:backbone-adaptation}

We investigate whether fine-tuning the TSFM improves forecasting after action conditioning has been learned.
The first stage trains \model\xspace with a frozen TimesFM~3.0 backbone.
Starting from the selected checkpoint, the second stage freezes the learned action-conditioning modules and either fine-tunes all TSFM parameters or tunes only the output head and rank-8 attention-layer LoRA.
Both variants minimize factual forecasting loss at a learning rate of $10^{-6}$ for up to ten epochs, stopping after three epochs without validation improvement.
As shown in \tabref{tab:backbone-adaptation}, both strategies slightly increase all three aggregate errors, with a larger degradation under full fine-tuning.
These results suggest that adapting pretrained TSFM parameters to action-dependent dynamics remains a nontrivial challenge, motivating fine-tuning strategies that better align temporal knowledge with action effects.

\begin{table}[h]
\centering
\caption{95\% confidence intervals (CIs) for $\Delta\mathrm{nRMSE}$ (method minus vanilla). Negative values indicate improvement.}
\label{tab:forecast-confidence}
\small
\setlength{\tabcolsep}{3pt}
\begin{tabular}{l|cccc}
\toprule
\textbf{Method} & \textbf{TimesFM} & \textbf{Chronos} & \textbf{Toto} & \textbf{Moirai} \\
\midrule
\textbf{TSFM++} & \makecell{$[-0.227,$\\$-0.010]$} & \makecell{$[-0.099,$\\$-0.001]$} & \makecell{$[-0.034,$\\$0.008]$} & \makecell{$[-0.089,$\\$-0.014]$} \\
\midrule
\textbf{\model} & \makecell{$[-0.427,$\\$-0.028]$} & \makecell{$[-0.124,$\\$-0.044]$} & \makecell{$[-0.246,$\\$-0.027]$} & \makecell{$[-0.243,$\\$-0.031]$} \\
\bottomrule
\end{tabular}
\end{table}

\section{Forecasting Uncertainty Analysis}
\label{app:forecast-confidence}

To assess whether forecasting gains persist under dataset resampling, we compute 95\% confidence intervals (CIs) for $\Delta\mathrm{nRMSE}$, defined as each method's overall nRMSE minus that of its corresponding vanilla backbone.
We perform 10,000 paired bootstrap resamples of the seven datasets, jointly sampling each dataset's history--horizon configurations and preserving their weights in the overall average.
The 2.5th and 97.5th percentiles define the interval bounds.
In \tabref{tab:forecast-confidence}, all four intervals for \model\xspace fall below zero, supporting lower nRMSE than vanilla forecasting, whereas the TSFM++ interval for Toto includes zero.
These results support the consistency of \model\xspace improvements over vanilla forecasting under dataset resampling across all four TSFMs.

\begin{table}[h]
\centering
\caption{Performance against prototype composition variant.}
\label{tab:prototype-composition}
\small
\setlength{\tabcolsep}{3pt}
\begin{tabular}{l|ccc}
\toprule
\textbf{Method} & \textbf{nMAE $\downarrow$} & \textbf{nRMSE $\downarrow$} & \textbf{MASE $\downarrow$} \\
\midrule
\textbf{\model} & \textbf{0.8997} & \textbf{1.0185} & \textbf{3.0651} \\
\textbf{Random reassignment} & 0.9215 & 1.0438 & 3.1586 \\
\bottomrule
\end{tabular}
\end{table}

\section{Prototype Composition Analysis}
\label{app:prototype-composition}

To assess the importance of input-dependent prototype composition, we compare \model\xspace with a random-reassignment variant of TimesFM~3.0 across datasets.
At each forward pass, this variant replaces every future patch's mixture weights with a vector sampled from a fixed bank of weights extracted by the original model on training instances.
We retrain its learnable modules from scratch using the original training protocol, with random reassignment applied throughout training, validation, and testing.
\tabref{tab:prototype-composition} shows that input-dependent composition achieves lower errors on all three metrics, supporting the value of matching prototype mixtures to the current input.

\begin{figure}[h]
\centering
\includegraphics[width=0.8\linewidth]{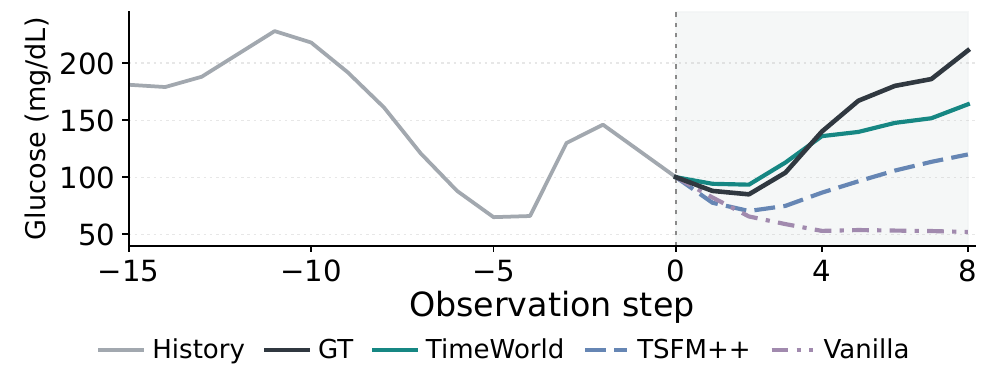}
\caption{Forecasting example on HUPA-UCM with Toto.}
\label{fig:case-action}
\end{figure}

\section{Case Study}
\label{app:case-study}

To examine how action information helps predict a departure from recent observations, we study a Toto-based forecast on HUPA-UCM in \figref{fig:case-action}.
The input comprises 96 glucose observations and a recorded event combining carbohydrate intake of 2.5 portions with 2 IU of rapid-acting insulin.
The next eight observations briefly decline before rising from 85 to 211 mg/dL, reversing the recent downward trajectory.
This reversal makes continuation of the historical trend inadequate and provides a concrete test of forecasting under the recorded action context.
Vanilla continues downward, while TSFM++ anticipates a recovery but substantially underestimates its magnitude.
\model\xspace follows the initial rise more closely and reduces MAE by 59.41\% relative to TSFM++, although it still underestimates the later increase.
This case shows that \model\xspace better captures the observed trend reversal under the recorded action conditions.

\end{document}